\documentclass[11pt]{article}
\PassOptionsToPackage{numbers, compress}{natbib}
\usepackage[utf8]{inputenc}
\usepackage{amsmath}
\usepackage{amsfonts}
\usepackage{mathrsfs}
\usepackage{amssymb}
\usepackage{amsthm}

\usepackage{libertine}
\usepackage{wrapfig}
\usepackage{graphicx}
\usepackage{multirow}
\usepackage{epstopdf}
\usepackage{multicol}
\usepackage{subfigure}
\usepackage{booktabs}

\usepackage{algorithmic}
\usepackage{algorithm}
\usepackage[algo2e,linesnumbered]{algorithm2e}
\usepackage{stackengine}
\usepackage{graphicx}
\usepackage{multirow}
\usepackage{epstopdf}
\usepackage{multicol}
\usepackage{caption}
\usepackage{subfigure}
\usepackage{enumitem}
\usepackage{amsmath}
\usepackage{pifont}
\usepackage{natbib}
\usepackage{bbm}
\usepackage{makecell}
\usepackage{mathtools}
\usepackage[table]{xcolor}
\definecolor{mydarkred}{rgb}{0.6,0,0}
\definecolor{mydarkgreen}{rgb}{0,0.6,0}
\usepackage[colorlinks,
linkcolor=mydarkred,
citecolor=mydarkgreen]{hyperref}
\usepackage{url}
\usepackage{bm}
\usepackage{pifont}
\usepackage{stfloats}
\usepackage[T1]{fontenc}
\usepackage{arydshln}
\usepackage{colortbl}
\usepackage{balance}

\usepackage{tikz}
\newcommand{\eg}{\textit{e.g.}}
\newcommand{\ie}{\textit{i.e.}}
\def\nopo{P(\tilde{\bm{Y}}|X = \bm{x})}
\def\clpo{P(\bm{Y}|X = \bm{x})}

\def\nopos{P(\tilde{\bm{Y}}|X)}
\def\clpos{P(\bm{Y}|X)}

\usetikzlibrary{positioning,shapes,arrows}
\title{Do We Need to Penalize Variance of Losses for Learning with Label Noise?}

\author{
  Yexiong Lin$^{1}$,
  Yu Yao$^{2}$,
  Yuxuan Du$^{2}$,\\
  Jun Yu$^{3}$,
    Bo Han$^{4}$, Mingming Gong$^{5}$, Tongliang Liu$^{\dagger\text{ }2}$\\[1ex]
  $^1$Hunan University;
  $^2$University of Sydney;\\
  $^3$University of Science and Technology of China;\\
  $^4$Hong Kong Baptist University;
  $^5$University of Melbourne;\\
}

\date{}
\begin{document}

\maketitle

\begin{abstract}
Algorithms which minimize the averaged loss have been widely designed for dealing with noisy labels. Intuitively, when there is a finite training sample, penalizing the variance of losses will improve the stability and generalization of the algorithms. Interestingly, we found that the variance should be increased for the problem of learning with noisy labels. Specifically, increasing the variance will boost the memorization effects and reduce the harmfulness of incorrect labels. By exploiting the label noise transition matrix, regularizers can be easily designed to reduce the variance of losses and be plugged in many existing algorithms. Empirically, the proposed method by increasing the variance of losses significantly improves the generalization ability of baselines on both synthetic and real-world datasets.\let\thefootnote\relax\footnotetext {$\dagger$ Correspondence author.}
\end{abstract}

\newpage
\section{Introduction}

Learning with noisy labels can be dated back to \cite{angluin1988learning}.
It has recently drawn a lot of attention \cite{liu2015classification, nguyen2019self,li2020dividemix,li2021provably} because large-scale datasets used in training modern deep learning models can easily contain label noise, \eg, ImageNet \cite{deng2009imagenet} and Clothing1M \cite{xiao2015learning}.
The reason is that it is expensive and sometimes infeasible to manually annotate large-scale datasets. Meanwhile, many cheap but imperfect surrogates such as crowdsourcing and web crawling are widely used to build large-scale datasets. Training with such data can lead to poor generalization abilities of modern deep learning models because they can memorize noisy labels \cite{han2018co,zhang2021understanding}.

Generally, the algorithms of learning with noisy labels can be divided into two categories: \textit{statistically inconsistent algorithms} and \textit{statistically consistent algorithms}.  Methods in the first category are heuristic, such as selecting reliable examples to train model \cite{han2020sigua, yao2020searching, yu2019does, han2018co, malach2017decoupling, ren2018learning, jiang2018mentornet}, correcting labels \cite{ma2018dimensionality, kremer2018robust, tanaka2018joint, reed2014training}, and adding regularization \cite{han2018masking, guo2018curriculumnet, veit2017learning, vahdat2017toward, li2017learning, li2020gradient, wu2020class2simi}. Those methods empirically perform well. However, it is not guaranteed that the classifiers learned from noisy data are statistically consistent and often need extensive hyper-parameter tuning on clean data.

To address this problem, many researchers explore algorithms in the second category. Those algorithms aim to learn \textit{statically consistent classifiers} \cite{liu2015classification, northcutt2017learning, goldberger2016training, patrini2017making, thekumparampil2018robustness, liu2020peer, xu2019l_dmi, xia2020part}. 
Specifically, their risk functions are specially designed to ensure that minimizing their expected risks on the noise domain is equivalent to minimizing the expected risk on the clean domain.
In practice, it is infeasible to calculate the expected risk. 
To approximate the expected risk, existing methods minimize their empirical risks, \ie, the averaged loss over the noisy training examples, which is an unbiased estimator to the expected risk \cite{xia2019anchor,li2021provably}.
However, when the number of examples is limited, the variance of the empirical risk could be high, which leads to a large estimation error.

\begin{figure*} [t] 
\centering
        % \subfigure[Accuracy of increasing variance of losses]{\label{fig:fig2a}\includegraphics[width=0.30\textwidth]{./pic/Figure2a.pdf}}
        % \!
        % \subfigure[Accuracy of original loss]{\label{fig:fig2b}\includegraphics[width=0.30\textwidth]{./pic/Figure2b.pdf}}
        % \!
        % \subfigure[Accuracy of increasing variance of losses]{\label{fig:fig2c}\includegraphics[width=0.30\textwidth]{./pic/Figure2c.pdf}}
        % \!\\
        \subfigure[Penalizing variance of losses]{\label{fig:fig2a}\includegraphics[width=0.32\textwidth]{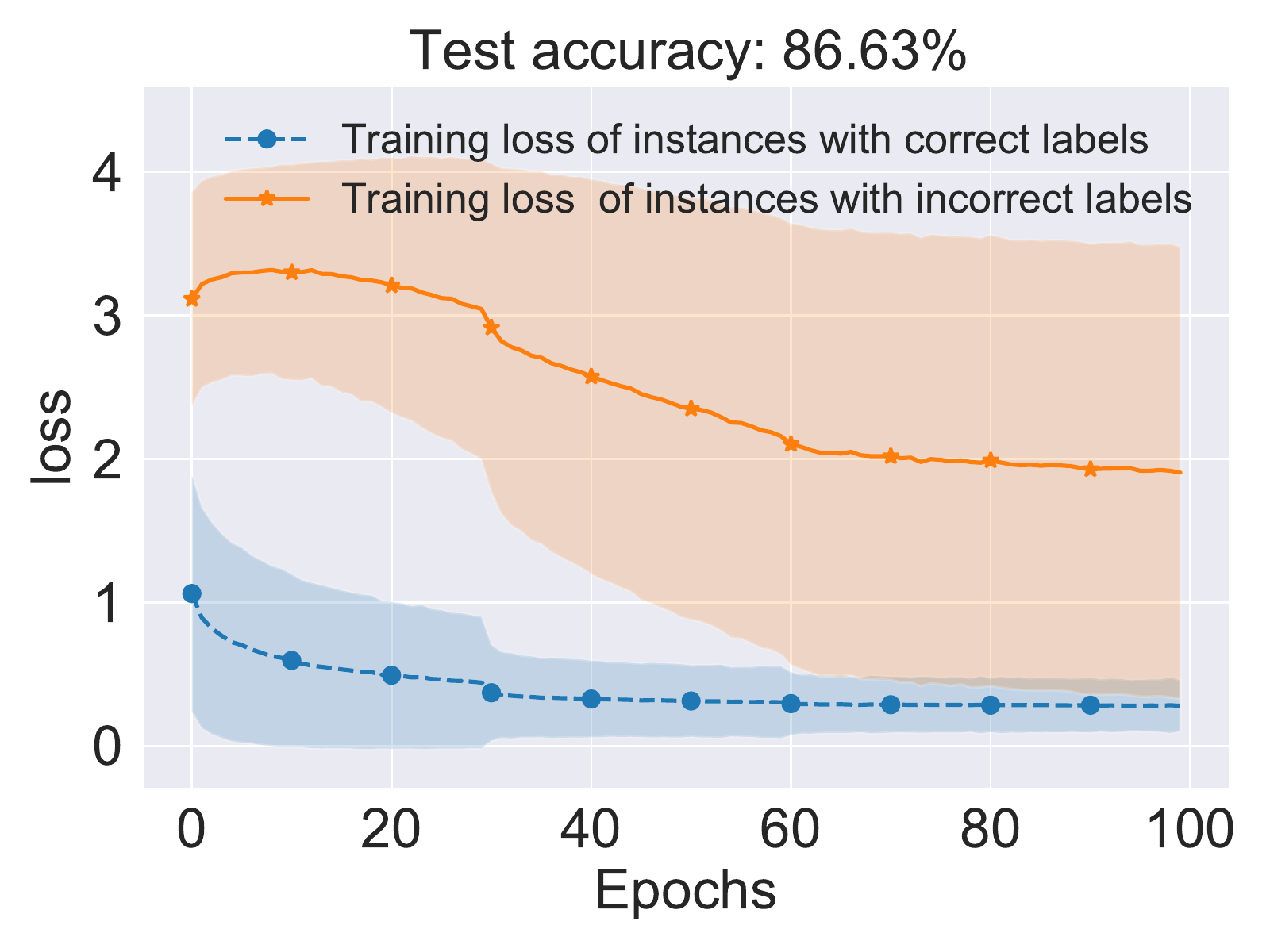}}
        \!
        \subfigure[Original loss]{\label{fig:fig2b}\includegraphics[width=0.32\textwidth]{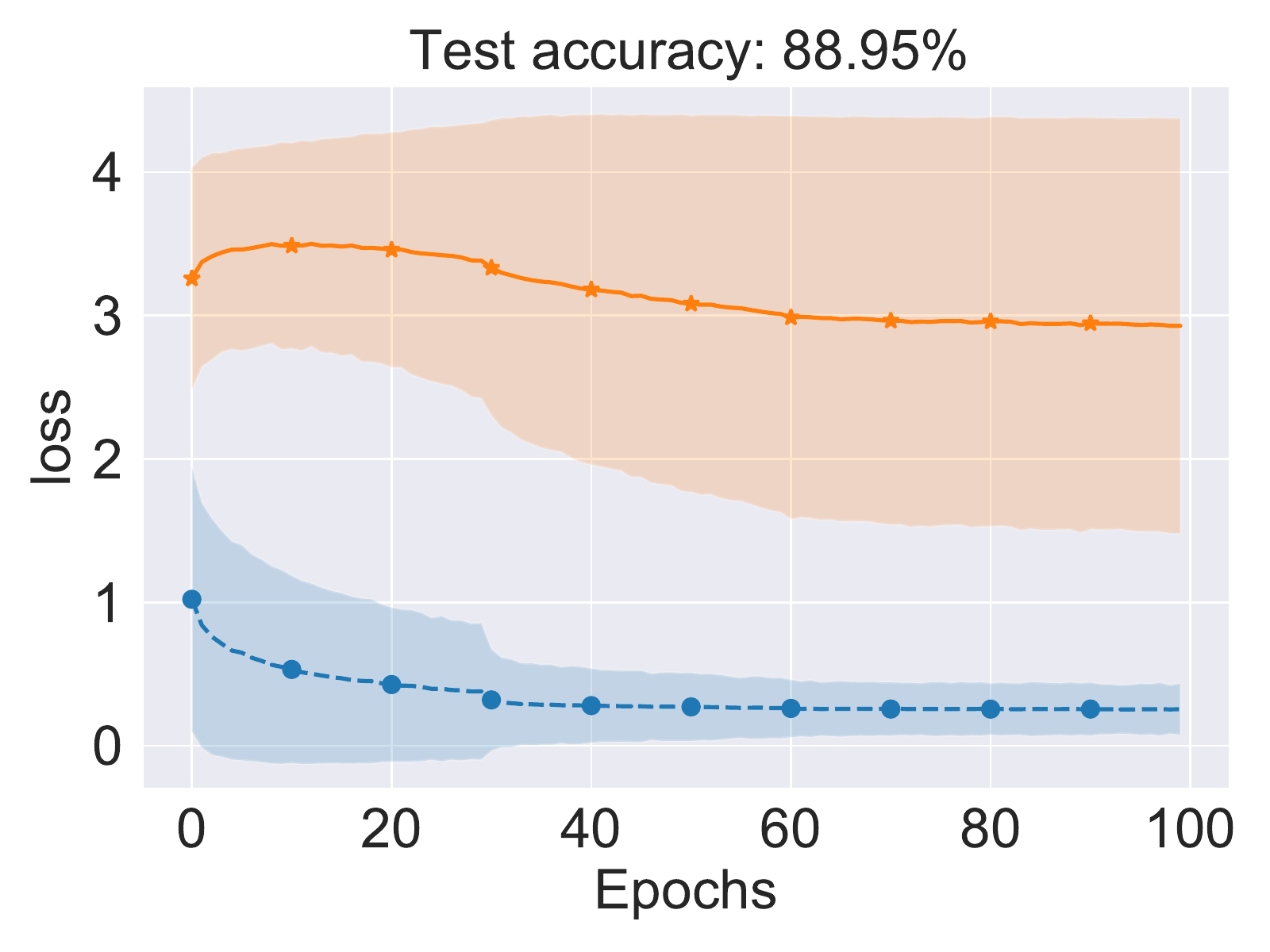}}
        \!
        \subfigure[Increasing variance of losses]{\label{fig:fig2c}\includegraphics[width=0.32\textwidth]{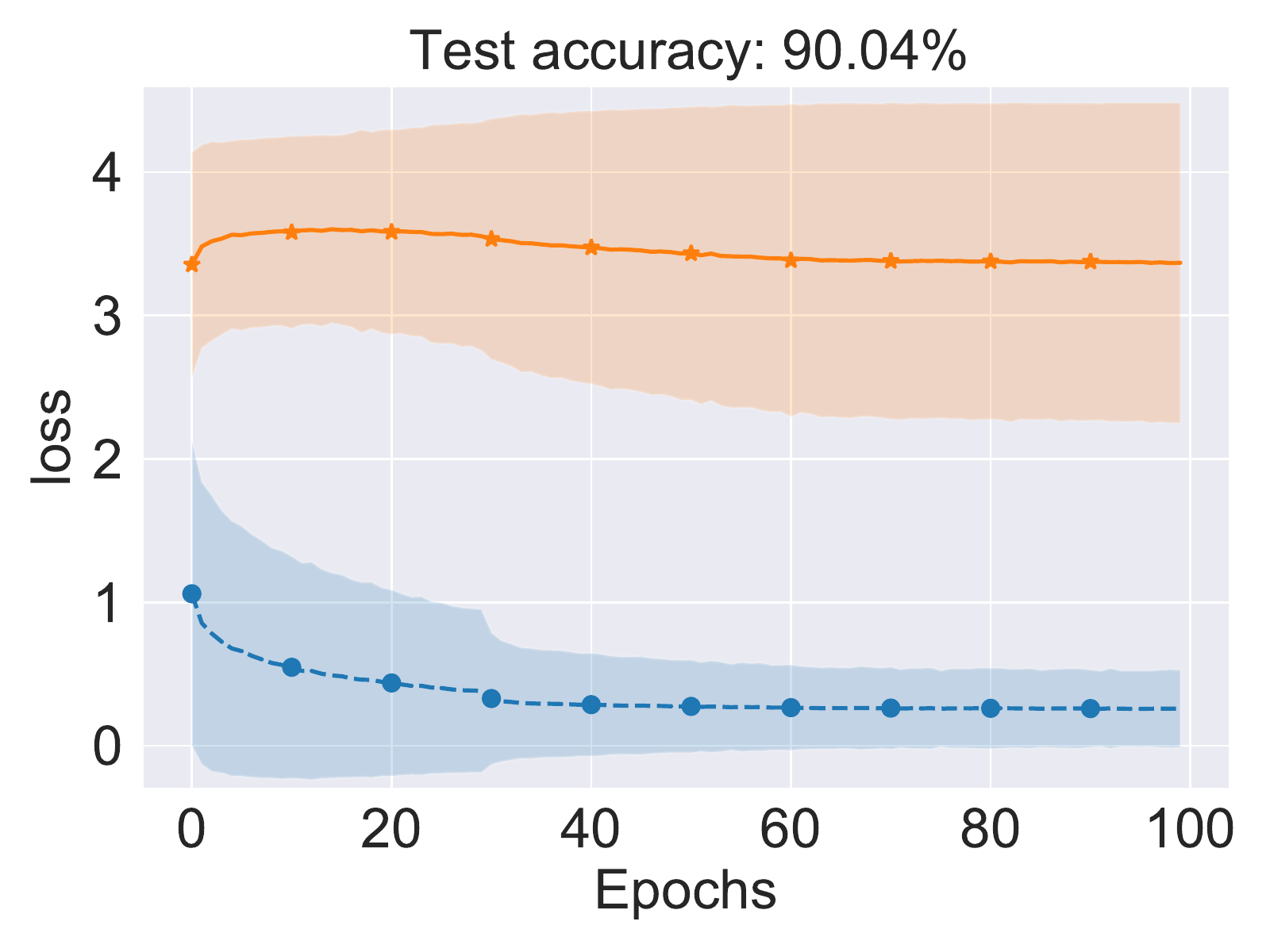}}
        \!
        \caption{We visualize the averaged training loss of instances with correct labels (blue dashed lines) and instances with incorrect labels (yellow solid lines) obtained by penalizing the variance, employing original loss, and increasing the variance in (a)-(c), respectively. The noise type is symmetry flipping, and the noise rate is $0.2$. The neural network ResNet-18 and the baseline Forward \cite{patrini2017making} are employed. The transition matrix $\bm{T}$ is given and does not need to be estimated.}\label{fig:fig2}
\end{figure*}

However, we report that penalizing the variance is generally not helpful for the problem of learning with noisy labels. By contrast, we should increase the variance, which will boost the memorization effects \cite{bai2021me} and reduce the harmfulness of incorrect labels.
% , so we proposed a simple but effective method to boost the model performance which is called LVR (Losses Variance Regularizer). 
This is because deep neural networks tend to learn easy samples first due to memorization effects \cite{bai2021me}, then loss of instances with incorrect labels are probably larger than those of correct instances \cite{han2018co}. Penalizing the variance of losses could force the model to reduce the loss of the instances with incorrect labels, then it will lead to performance degeneration. 
In contrast, increasing the variance of losses could efficiently prevent the model from overfitting noisy labels of examples with large losses. In Section 3, we further show that increasing the variance of losses can be seen as a weighting method which assigns small weights to the gradients of large losses and large weights to the gradients of small losses. 

% Compared (a), (b) and (c), change of the variance does not have much influence on the averaged training loss of instances with correct labels, but the averaged training loss of instances with incorrect labels is very different. Penalizing the variance leads to the averaged training loss of instances with incorrect labels decreasing fast. It encourages the model to overfit instances with incorrect labels. On the contrary, increasing variance can prevent the averaged training loss of instances with incorrect labels from decreasing.

Intuitively, as illustrated in Fig. \ref{fig:fig2}, change of the variance does not have much influence on the averaged training loss of instances with correct labels, but the averaged training loss of instances with incorrect labels is very different.  Specifically, penalizing the variance leads to the averaged training loss of instances with incorrect labels decreasing fast, which will encourage the model to overfit instances with incorrect labels. On the contrary, increasing variance can prevent the averaged training loss of instances with incorrect labels from decreasing as shown in Fig. \ref{fig:fig2c}. Therefore, the memorization effects are boosted. As a result, the test accuracy is improved significantly by encouraging the variance of losses.

The rest of this paper is organized as follows. In Section 2, we introduce related work. In Section 3, we propose our method and its advantages. Experimental results on both synthetic and real-world datasets are provided in Section 4. Finally, we conclude the paper in Section 5.

\section{Related Works}

Some methods proposed to reduce the side-effect of noisy labels using heuristics, For example, many methods utilize the memorization effects to select reliable examples \cite{han2020sigua, yao2020searching, yu2019does, malach2017decoupling, ren2018learning, jiang2018mentornet} or correcting labels \cite{ma2018dimensionality, kremer2018robust, tanaka2018joint, reed2014training}. Those methods empirically perform well. However, most of them do not provide statistical guarantees for learned classifiers on noisy data. Some methods treat noisy labels as outliers and focus on designing bounded
loss functions \cite{ghosh2017robust,gong2018decomposition, wang2019symmetric,shu2020learning}. For example, a symmetric cross-entropy loss has been proposed which has proven to be robust to label noise \cite{wang2019symmetric}.
These methods focus on the numerical property of loss functions, and the designed loss function could be proved to be noise-tolerant if the noise rate is not large. 

To be more noise-tolerant, some methods aim to build the \textit{consistent risk} \cite{liu2015classification, scott2015rate, goldberger2016training, northcutt2017learning, thekumparampil2018robustness, xia2020part, li2021provably}. 
Specifically, their risk functions are specially designed to ensure that minimizing their expected risks on the noise domain is equivalent to minimizing the expected risk on the clean domain. 
To achieve this, the noise transition matrix $\bm{T}(x)\in [0,1]^{C \times C}$ usually has to be identified \cite{patrini2017making,li2021provably}, where $C$ is the number of classes. 
% When the noise is independent of instance, \ie, $\bm{T}(x) = \bm{T}$, the transition matrix $\bm{T}$ can be learned by discovering anchor points \cite{liu2015classification, patrini2017making, xia2019anchor}, exploiting clusterability \cite{zhu2021clusterability}, regularizing total variation \cite{zhang2021learning}, or minimizing volume of the matrix $\bm{T}$ \cite{li2021provably}.
% Then the clean class posterior $P(\tilde{Y}|X=x)$ can be inferred by using the transition matrix $\bm{T}$ and the noisy class posterior $P(Y|X=x)
An example is that the clean class posterior $\clpo :=[P(Y=1|X=\bm{x}),\ldots, P(Y=C|X=\bm{x})]^\top$ can be inferred by utilizing the noisy class posterior $\nopo$ and the transition matrix $\bm{T}(\bm{x})$, where $\bm{T}_{ij}(\bm{x}) = P(\tilde{Y}=i | Y=j, X )$, \ie, $\clpo = [\bm{T}(\bm{x})]^{-1} \nopo$. 
Then, the expected risk of a function $f(X,Y)$ modeling $\clpos$ can be formulated as the expected risk of a function $g(X,\tilde{Y})$ modeling $\nopos$ multiplied by $\bm{T}(X)$, \ie,
\begin{align*}
    R(f(X,Y)) = R([\bm{T}(X)]^{-1}g(X,\tilde{Y})).
\end{align*}
In practice, the expected risk $R([\bm{T}(X)]^{-1}g(X,\tilde{Y}))$ is infeasible to calculate, existing methods approximate the expected risk with the averaged loss over the noisy training examples.
Normally, when the number of samples is limited, the variance of the empirical risk could be high, which could lead to a large estimation error.

To reduce the estimation error and improve the stability, different variance reduction methods have been proposed. 
For example, Truncated Importance Sampling \cite{ionides2008truncated} limits the maximum importance weight, which decreases the mean squared estimation error of the standard importance sampling.
\citet{anschel2017averaged} proposed to stable training procedure and improve performance by reducing approximation error variance in target rewards.
\citet{achab2015sgd} proposed to use stochastic gradient descent (SGD) with variance reduction for optimizing a finite average of smooth convex functions, and a linear rate of convergence under strong convexity is proved. Similarly, \citet{allen2016variance} proved that a fast convergence rate can be achieved by using variance reduction on the non-convex optimization problem.

\section{Variance of Losses Enhancement for Learning with Label Noise}
\label{sec:var}
%\section{Per-Instance-variance of losses Enhancement for Learning with Label Noise}
In this section, we propose our method losses Variance Regularization for label-Noise Learning (VRNL). We reveal how the proposed method reduces the negative effects of noisy labels. We also illustrate the advantage of our regularizer when working with existing methods.

\subsection{Methodology}
We show that the proposed method can efficiently prevent the model from learning incorrect labels.
Intuitively, by encouraging the variance of losses, the instances with incorrect labels will be suppressed, and instances with correct labels will be magnified.
Theoretically, the gradients of large-loss examples will be assigned with small weights, and gradients of small-loss examples will be assigned with large weights.
As a result, the model puts more trust on small-loss examples; and small-loss examples will contribute more to the update of parameters, which could boost the memorization effects of deep networks.

To analyze the regularization effects in our method, we have to define some notations here. Let $Var[.]$ denote variance of a distribution and $C$ denote the number of classes. Let $f_{\theta}:\mathcal{X}\to \Delta^{C-1}$ be a mapping parameterized by $\theta$ (\eg, a neural network), where $\Delta^{C-1}$ denotes a probability simplex. Generally, the expected risk of a label-noise learning method is formulated as $\mathbb{E}_{(X, \tilde{Y})}[\ell(f_\theta(X),\tilde{Y})]$, where $\ell(\cdot)$ is the loss function of the method. 
We proposed to add a variance regularizer to the expected risk. Specifically, the objective function of our method is 
\begin{align}
\label{eq:general}
R_G(f_\theta)=\mathbb{E}_{(X, \tilde{Y})}[\ell(f_\theta(X),\tilde{Y})] -\alpha Var_{(X, \tilde{Y})}[\ell(f_\theta(X),\tilde{Y})],
% &=\frac{1}{n}\sum_{i=1}^{n}[\ell(f_\theta(X),\tilde{Y})]\nonumber \\
% &-\alpha\left(\frac{1}{n}\sum_{i=1}^{n}[\ell(f_\theta(X),\tilde{Y})^2]\nonumber \right. \\
% &\left. -\mathbb{E}^2_{(X,\tilde{Y}) \sim D_{\rho}}[\ell(f_\theta(X),\tilde{Y})]\right),
\end{align}
where $Var_{(X, \tilde{Y})}[\ell(f_\theta(X),\tilde{Y})]$ is a regularization term, and $\alpha$ is an adjustable hyper-parameter to control the strength of the regularization effects. 
To encourage the variance of losses, $\alpha$ is chosen to be a positive value. Usually, the strength of the regularization effects should not be too large, and $\alpha$ is much smaller than $1$.

To exploit the influence of our regularizer to the update of parameter $\theta$, we first illustrate the derivative of the objective function to $\theta$, i.e.,

\begin{align}
\label{eq:gradient}
\frac{R_G(f_\theta)}{\partial \theta}&=\frac{\partial \mathbb{E}_{(X, \tilde{Y})}[\ell(f_{\theta}(X),\tilde{Y})]}{\partial \theta}-\alpha \frac{\partial Var_{(X,\tilde{Y})}[\ell(f_\theta(X),\tilde{Y})]}{{\partial \theta}}\nonumber \\
% &=\mathbb{E}_{(x, \tilde{y})\sim D_{\rho}}[\frac{\partial \ell(f_{\theta}(X),\tilde{Y})}{\partial \theta}]\nonumber\\
% &-\alpha \left \{\mathbb{E}_{(x, \tilde{y})\sim D_{\rho}}\left[2\ell(f_{\theta}(X),\tilde{Y})\frac{\partial \ell(f_{\theta}(X),\tilde{Y})}{\partial \theta}\right]\right.\nonumber\\
% &\left.-2\mathbb{E}_{(x, \tilde{y})\sim D_{\rho}}[\ell(f_{\theta}(X),\tilde{Y})]\mathbb{E}_{(x, \tilde{y})\sim D_{\rho}}\left [\frac{\partial \ell(f_{\theta}(X),\tilde{Y})}{\partial \theta}\right]\right\}\nonumber\\
% &= \mathbb{E}_{(x, \tilde{y})\sim D_{\rho}}\left[\left(1+2\alpha \mathbb{E}_{(x, \tilde{y})\sim D_{\rho}}[\ell(f_{\theta}(X),\tilde{Y})] \right. \right. \nonumber\\
% &\left.  -2 \alpha \ell(f_{\theta}(X),\tilde{Y})\left) \frac{\partial \ell(f_{\theta}(X),\tilde{Y})}{\partial \theta}\right. \right] \nonumber\\
&=\mathbb{E}_{(X, \tilde{Y})}\left[ W\frac{\partial \ell(f_{\theta}(X),\tilde{Y})}{\partial \theta}\right],
\end{align}
where $W$ is a random variable, which is defined to be 
$$W=1+2\alpha \left(\mathbb{E}_{(X, \tilde{Y})}[\ell(f_{\theta}(X),\tilde{Y})]- \ell(f_{\theta}(X),\tilde{Y})\right).$$

For a specific example $(x, \tilde{y})$, its corresponding gradient is $w\frac{\partial \ell(f_{\theta}(x),\tilde{y})}{\partial \theta}$,
where its weight $w$ is 
$$w=1+2\alpha \left(\mathbb{E}_{(X, \tilde{Y})}[\ell(f_{\theta}(X),\tilde{Y})]- \ell(f_{\theta}(x),\tilde{y})\right).$$

As aforementioned, $\alpha$ is chosen to be small enough such that $w$ should be positive.
The above equation shows that 1). if the loss of the example $(x,\tilde{y})$ is smaller than the mean of the losses of noisy distribution, $\left(\mathbb{E}_{(X, \tilde{Y})}[\ell(f_{\theta}(X),\tilde{Y})]- \ell(f_{\theta}(x),\tilde{y})\right)$ will be positive, and the weight associated with its gradient is larger than $1$. Then the example contributes more to the update of parameter $\theta$.
2). If the loss of the example $(x,\tilde{y})$ is larger than the mean loss of noisy distribution, $\left(\mathbb{E}_{(X, \tilde{Y})}[\ell(f_{\theta}(X),\tilde{Y})]- \ell(f_{\theta}(x),\tilde{y})\right)$ will be negative. The weight associated with its gradient is small. 
%, meanwhile the $\alpha$ which is the coefficient before the regularization term is small enough, so the weight $w$ is positive. 
Then the example contributes less to the update of parameter $\theta$. 

% \footnote{Here we assume that the loss function $\ell$ is bounded, and $w$ are always positive. Empirically, this can be easily satisfied. For example, by adding a small value such as $1e-7$ to cross-entropy loss, and let hyper-parameter $\alpha$ be small. }.\footnote{learn how to use footnote.}

Due to the memorization effects, deep neural networks tend to learn easy instances first and gradually learn hard instances \cite{han2018co,arpit2017closer}. In learning with noisy labels, hard instances are more likely to have incorrect labels and should not be trusted \cite{bai2021understanding}. 
By employing the proposed method, the gradient of noisy examples is assigned with small weights. In such a way, noise examples would have less contribution to update of parameter $\theta$, which prevents the model from memorizing incorrect labels. 

% According to the memorization effects, neural networks tend to memorize the majority (clean) samples first and then overfit minority (noisy) samples gradually \cite{bai2021understanding}. Memorizing the noisy samples is harmful for performance, even if the loss are correct \cite{patrini2017making} or reweghting \cite{liu2015classification}. 

% By employing the proposed method, the gradient of noisy examples is assigned with small weights. Therefore, the model are prevent from memorizing the incorrect labels. As a result, our method can boost the model perfermance dramatically. 

Additionally, compared with existing small-loss based methods, our method can sufficiently exploit the information contained in the whole training dataset. Existing learning with noisy labels methods \cite{han2018co,nguyen2019self} usually divide the training sample into confident examples and unconfident examples based on the small-loss trick \cite{jiang2018mentornet,malach2017decoupling,li2020dividemix}. To be specific, the examples with large losses are unconfident examples, and their labels are ignored. However, some of the unconfident examples are hard-clean examples that contain useful information for learning noise-robust classifiers \cite{bai2021me}. 
In contrast, our method does not ignore unconfident examples but assign them with small weights such that all the label information of the training dataset is carefully exploited.

% Secondly, the proportion of the confident examples selected from a training sample is usually determined by a hyper-parameter $\tau$ \cite{malach2017decoupling,han2018co, li2020dividemix}. In different scenarios, the value of $\tau$ has to be carefully selected.
% In contrast, for our method, the hyper-parameter $\tau$ is replaced by an interpretable weight $w$ which can be directly calculated according to the deviation of each loss from the mean loss.
% Additionally, our method is general and can be easily designed to reduce the variance of losses and be plugged in many existing algorithms \cite{liu2015classification,patrini2017making,yao2020dual,li2021provably}.

\subsection{Practical Implementation}

In practice, the expected risk $R_G(f_\theta)$ in Eq.~\eqref{eq:general} can not be calculated, the empirical risk is employed as an unbiased approximation. Let $n$ be the number of training examples, generally, the empirical risk of our method is as follows:
\begin{align}\label{eq:practical}
    \hat{R}(f_\theta) &= \frac{1}{n}\sum_{i=1}^{n}  \ell(f_\theta(x_i),\tilde{y}_i)-\alpha\left(\frac{1}{n}\sum_{i=1}^{n}\ell(f_\theta(x_i),\tilde{y}_i)^2 \right.-\left.\left(\frac{1}{N}\sum_{i=1}^{n} \ell(f_\theta(x_i),\tilde{y}_i)\right)^2 \right).
\end{align}

We further illustrate specific forms and settings of our regularization when work with existing methods, \ie, Importance Reweighting \cite{liu2015classification}, Forward \cite{patrini2017making}, and VolMinNet \cite{li2021provably}. Empirically, our method improves their classification accuracy.

\textbf{Work with Forward.} Forward correction \cite{patrini2017making} exploits the noise transition matrix $\bm{T}$ to estimate clean class posterior distribution. We use the same methods in the original paper \cite{patrini2017making} to estimate the transition matrix.
%To estimate the transition matrix, we follow the original paper \cite{patrini2017making} \footnote{needs citation} which uses T-estimator\footnote{why mention this specific percentile? This only holds true for a specific dataset.\textcolor{green}{\checkmark} bad writting, cannot understand.} in place of the largest noisy class-posteriors to estimate the transition matrix. 
The objective loss function by combining our method with Forward can be formulated as follows:

\begin{align*}
    \hat{R}_{\text{Forward}}(\theta,\hat{\bm{T}})&=\frac{1}{n}\sum_{i=1}^n \ell_{CE}(\hat{\bm{T}}f_\theta(x_i),\tilde{y}_i)\\
&-\alpha\left(\frac{1}{n}\sum_{i=1}^{n}\ell_{CE}(\hat{\bm{T}}f_\theta(x_i),\tilde{y}_i)^2 \right.-\left.\left(\frac{1}{n}\sum_{i=1}^{n} \ell_{CE}(\hat{\bm{T}}f_\theta(x_i),\tilde{y}_i)\right)^2 \right),
\end{align*}
where $\ell_{CE}$ is the cross-entropy loss, $\hat{\bm{T}}$ is the estimated transition matrix, $f_\theta$ models the clean class-posterior distribution, $\hat{\bm{T}}f_\theta$ models the noisy class-posterior distribution.

\textbf{Work with Importance Reweighting.} Importance Reweighting uses the weighted empirical risk to estimate the empirical risk with respect to clean class-posterior distribution \cite{liu2015classification}. 
To calculate the weight of the empirical risk, both noisy class-posterior distribution and clean class-posterior distribution need to be estimated. The objective loss function by combining our method with Important Reweighting can be formulated as follows:

\begin{align}
\label{eq:ir_general}
\hat{R}_{IR}(f_\theta)&=\frac{1}{n}\sum_{i=1}^{n}\hat{\beta}_i \ell_{CE}(f_\theta(x_i),\tilde{y}_i) \nonumber \\
&-\alpha \left(\left(\sum_{i=1}^{n}\hat{\beta}_i \ell_{CE}(f_\theta(x_i),\tilde{y}_i)\right)^2\right. \left. -\sum_{i=1}^{n}\hat{\beta}_i^2 \ell_{CE}(f_\theta(x_i),\tilde{y}_i)^2\right),
% &=\frac{1}{n}\sum_{i=1}^{n}[\hat{\beta}_i \ell_{CE}(f_\theta(x),\tilde{y})]\nonumber \\
% &-\alpha \left(\frac{1}{n}\sum_{i=1}^{n}[\hat{\beta}_i \ell_{CE}(f_\theta(x),\tilde{y})^2]\nonumber \right. \\
% &\left. -\mathbb{E}^2_{(x,\tilde{y}) \sim D_{\rho}}[\hat{\beta}_i \ell_{CE}(f_\theta(x),\tilde{y})] \right),
\end{align}
where $\hat{\beta}_i=\frac{\hat{P}_D(y_i|x_i)}{\hat{P}_{D_{\rho}}(\tilde{y}_i|x_i)}$. 
%If the label is incorrect, $\hat{\beta}_i$ probably is small, and if the label is correct, $\hat{\beta}_i$ probably is large \cite{liu2015classification}. 
The gradient with respect to $\hat{R}_{IR}$ an example $(x_i,\tilde{y}_i)$ is as follows: 

%\vspace{-2mm}
\begin{align}
\label{eq:ir_gradient}
\nabla \hat{R}_{IR}(f_\theta,(x,\tilde{y}))
&=\hat{w}_i\left(\ell_{CE}(f_{\theta}(x_i),\tilde{y}_i)\frac{\partial \hat{\beta}_i}{\partial \theta} \right .+ \left . \hat{\beta}_i \frac{\partial \ell_{CE}(f_{\theta}(x_i),\tilde{y}_i)}{\partial \theta}\right),
\end{align}
% \begin{align*}
% w_i^1&=\ell_{CE}(f_{\theta}(x_i),\tilde{y}_i)\hat{w}_i
% \end{align*}
% and
% \begin{align*}
% w_2^i&=\hat{\beta}_iw_{reweight}^i,
% \end{align*}
%\vspace{-2mm}
where 
%\vspace{-2mm}
\begin{align*}
\hat{w}_i=1+2\alpha \left(\frac{1}{n}\sum_{j=1}^{n}\hat{\beta}_j \ell_{CE}(f_{\theta}(x_j),\tilde{y}_j) \right.\left. -\hat{\beta}_i\ell_{CE}(f_{\theta}(x_i),\tilde{y}_i)\right).
\end{align*}

The $\ell_{CE}(f_{\theta}(x),\tilde{y})\frac{\partial \hat{\beta}_i}{\partial \theta} \nonumber+ \hat{\beta}_i \frac{\partial \ell_{CE}(f_{\theta}(x),\tilde{y})}{\partial \theta}$ is the gradient of the original Importance Reweighting loss. When the label $\tilde{y}_i$ is incorrect, the Reweighted loss $\hat{\beta}_i\ell_{CE}(f_{\theta}(x_i),\tilde{y}_i)$ is usually larger than the averaged loss $\frac{1}{n}\sum_{j=1}^{n}\hat{\beta}_j \ell_{CE}(f_{\theta}(x_j),\tilde{y}_j)$. Then their difference is negative, which lead the weight $\hat{w}_i$ to be small because the hyper-parameter $\alpha$ is positive. 
%At the same time, $\hat{\beta}_i$ is usually small, 
As a result, the instance with an incorrect label has a small contribution to the update of parameter $\theta$, the proposed method can prevent the model from memorizing the incorrect labels.

In the implementation, the early stopping technique is used for the approximation of the clean class-posterior distribution. Specifically, the model $f_{\theta}$ is trained on noisy data with $20$ epochs, and we feed the model output to a softmax function, then use the output of the softmax function $g(x)$ to approximate the clean class-posterior distribution. The noise transition matrix $\bm{T}$ has also been estimated by using the same approach as in Forward correction. Then the model $f_{\theta}$ is further optimized by both weighted empirical risk and regularization for the variance of losses as follows:

% \begin{align}
% \label{eq:ir_implement}
% \hat{R}_{ir}(\theta)&=\frac{1}{n}\sum_{i=1}^n\left[\ell_{CE}(f_\theta(x_i),\tilde{y}_i) \frac{g_{\tilde{y}}(x_i)}{(\hat{\bm{T}}g)_{\tilde{y}}(x_i)}\right] \nonumber \\
% &-\alpha \left(\frac{1}{n}\sum_{i}^{n}\left(\ell_{CE}(f_\theta(x_i),\tilde{y}_i)\frac{g_{\tilde{y}}(x_i)}{(\hat{\bm{T}}g)_{\tilde{y}}(x_i)}\right)^2 \right. \nonumber \\
% &-\left.\left(\frac{1}{n}\sum_{i}^{n} \ell_{CE}(f_\theta(x_i),\tilde{y}_i)\frac{g_{\tilde{y}}(x_i)}{(\hat{\bm{T}}g)_{\tilde{y}}(x_i)}\right)^2 \right).
% \end{align}

\begin{align}
\label{eq:ir_implement}
\hat{R}_{ir}(\theta)&=\frac{1}{n}\sum_{i=1}^n\left[\ell_{CE}(f_\theta(x_i),\tilde{y}_i) \frac{g_{\tilde{y}}(x_i)}{(\hat{\bm{T}}g)_{\tilde{y}}(x_i)}\right]-\alpha \hat{\sigma}^2_{\theta},
\end{align}
where
\begin{align}
\label{eq:ir_implement}
\hat{\sigma}^2_{\theta}&= \left(\frac{1}{n}\sum_{i=1}^{n}\left(\ell_{CE}(f_\theta(x_i),\tilde{y}_i)\frac{g_{\tilde{y}}(x_i)}{(\hat{\bm{T}}g)_{\tilde{y}}(x_i)}\right)^2 \right.&-\left.\left(\frac{1}{n}\sum_{i=1}^{n} \ell_{CE}(f_\theta(x_i),\tilde{y}_i)\frac{g_{\tilde{y}}(x_i)}{(\hat{\bm{T}}g)_{\tilde{y}}(x_i)}\right)^2 \right).\nonumber
\end{align}

\textbf{Work with VolMinNet.} VolMinNet is an end-to-end label-noise learning method that learns the transition matrix and the clean class-posterior distribution simultaneously \cite{li2021provably}. It optimizes two objectives: 1). a trainable diagonally dominant column stochastic matrix $\hat{\bm{T}}$ by minimizing the log determinate $\log \det(\hat{\bm{T}})$; 2). the parameter $\theta$ of the model by the cross-entropy loss between the noisy label and the predicted probability by the neural network. In experiments, our VRNL only regularizes the parameter $\theta$ by calculating the variance of cross-entropy losses. The objective by combining our method with VolMinNet can be formulated as follows:

\begin{align*}
\hat{R}_{vol}(\theta,\hat{\bm{T}})&=\frac{1}{n}\sum_{i=1}^n \ell_{CE}(\hat{\bm{T}}f_\theta(x_i),\tilde{y}_i)+\lambda \log \det(\hat{\bm{T}})\\
&-\alpha\left(\frac{1}{n}\sum_{i=1}^{n}\ell_{CE}(\hat{\bm{T}}f_\theta(x_i),\tilde{y}_i)^2 \right. -\left.\left(\frac{1}{n}\sum_{i=1}^{n} \ell_{CE}(\hat{\bm{T}}f_\theta(x_i),\tilde{y}_i)\right)^2 \right),
\end{align*}

where $\lambda>0$ is an adjustable hyper-parameter, we set $\lambda=0.0001$ in all experiments. The transition matrix $\hat{\bm{T}}$ should be differentiable, diagonally dominant and column stochastic.

% \textbf{Help learn the Clean Class-posterior Distribution.} 
% \begin{figure}[t]
%     \centering
%     \includegraphics[width=0.45\textwidth,trim=20 20 40 36,clip]{Figure3.pdf}
%     \caption{The boxplot of confidence scores. We trained the ResNet-18 model on CIFAR10 under symmetric noise, noise rate is 50\%, and seleted the best model according to validation accuracy. We fed all samples of trainset, then get the noisy class posterior according to Equation \ref{eq:T}, and get the max value of corresponding training labels as confidence scores. The green lines are mean value, and the orange lines are median value. The mean value of noisy class-posteriors trained with variance regularization is smaller than trained without variance regularization one.}
%     \label{fig:conf}
% \end{figure}

Our method could help the state-of-the-art method VolMinNet \cite{li2021provably} to better estimate the transition matrix and the clean class-posterior distribution. 
Specifically, VolMinNet requires the estimated class-posteriors to be diverse, which is called the \textit{sufficiently scattered} assumption \cite{li2021provably}.
By increasing the variance of the loss, the diversity of the estimated noisy class posteriors is encouraged, so the estimated clean class-posteriors are also encouraged. Then transition matrix can be better learned, which leads to the clean class-posterior distribution being better estimated.

\begin{figure*} [t] 
\centering
% \captionsetup[subfigure]{labelformat=empty}
\subfigure[]{\label{fig:ir_clean2}\includegraphics[width=0.245\textwidth]{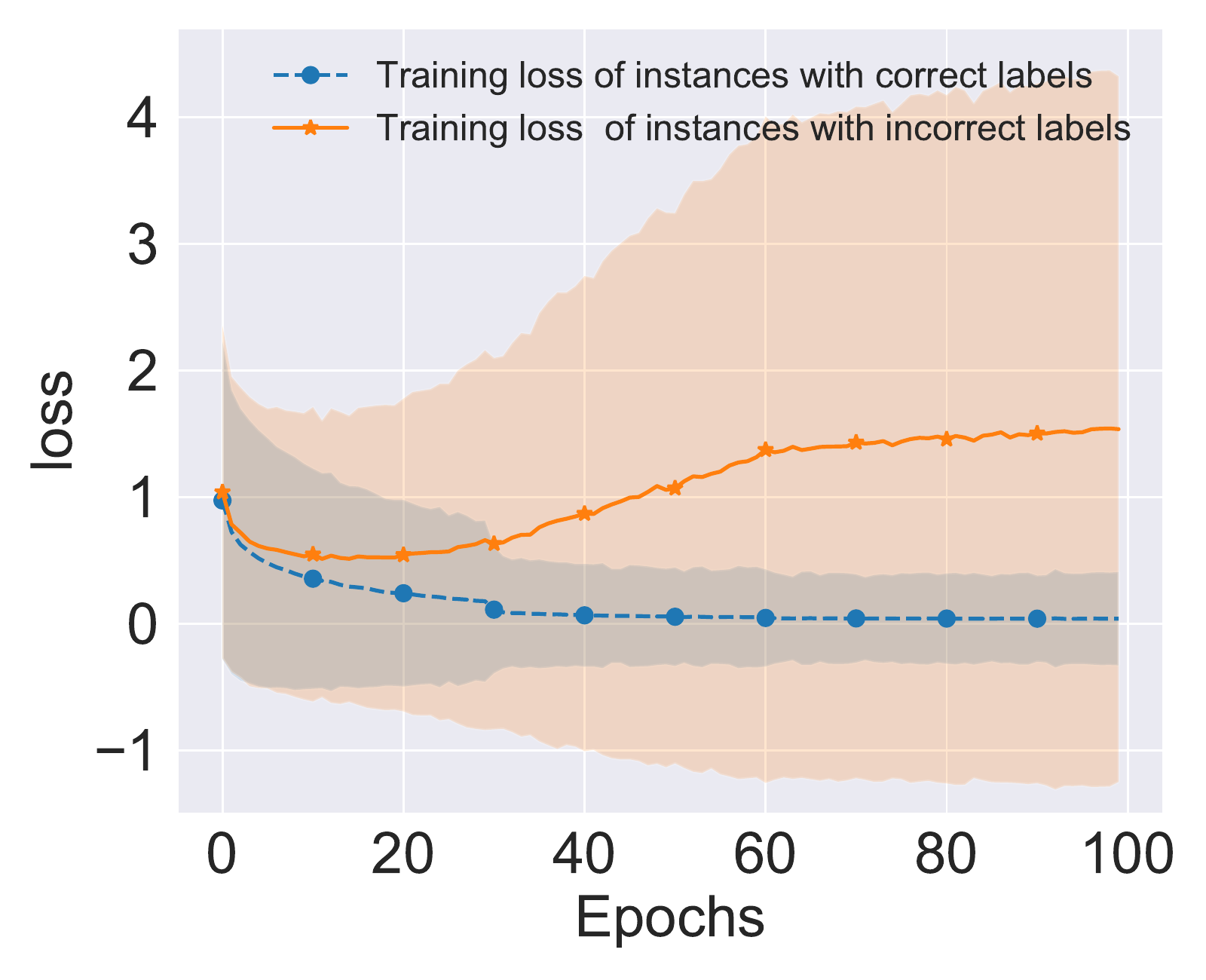}}
\subfigure[]{\label{fig:ir_clean3}\includegraphics[width=0.245\textwidth]{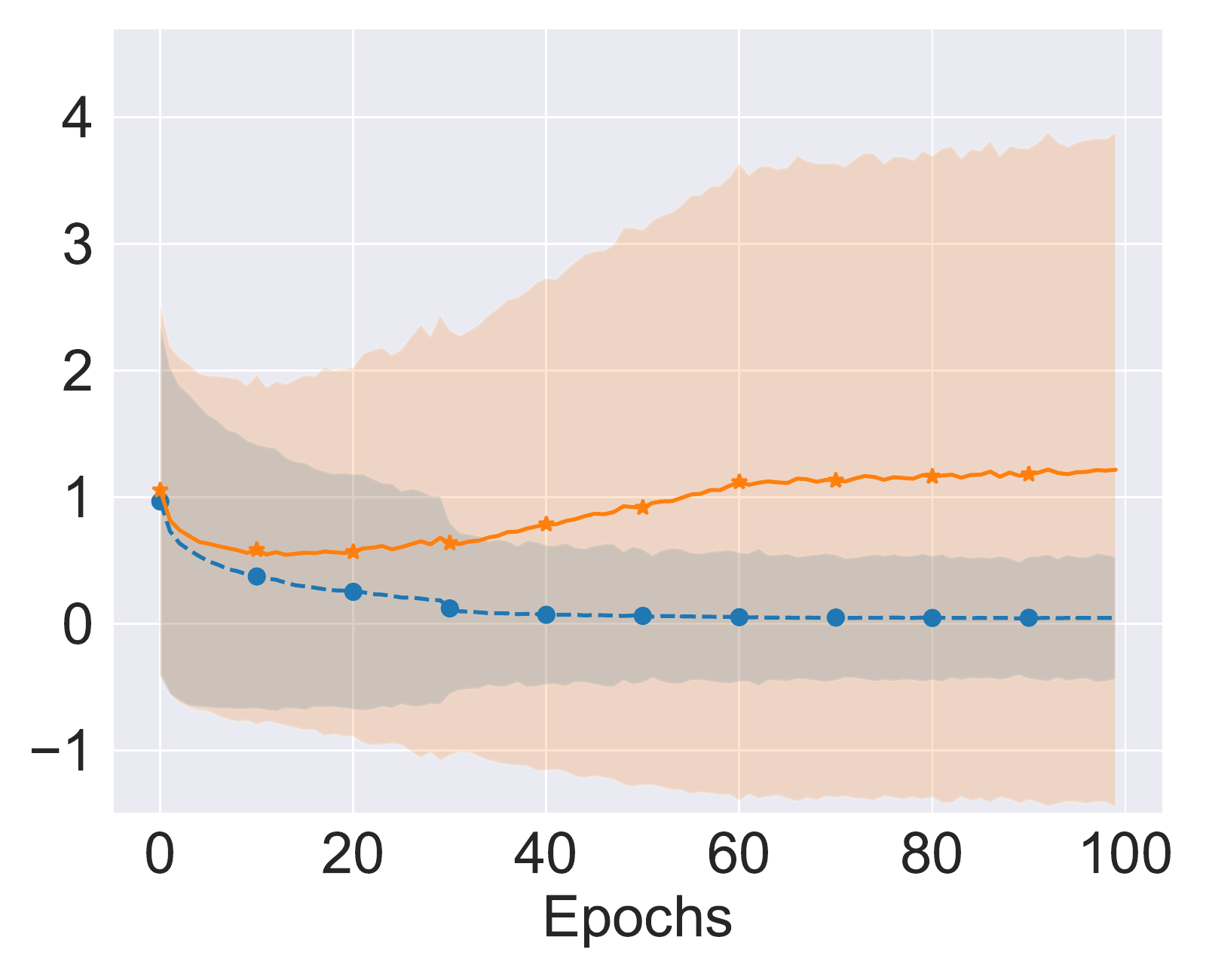}}
\subfigure[]{\label{fig:ce2}\includegraphics[width=0.245\textwidth]{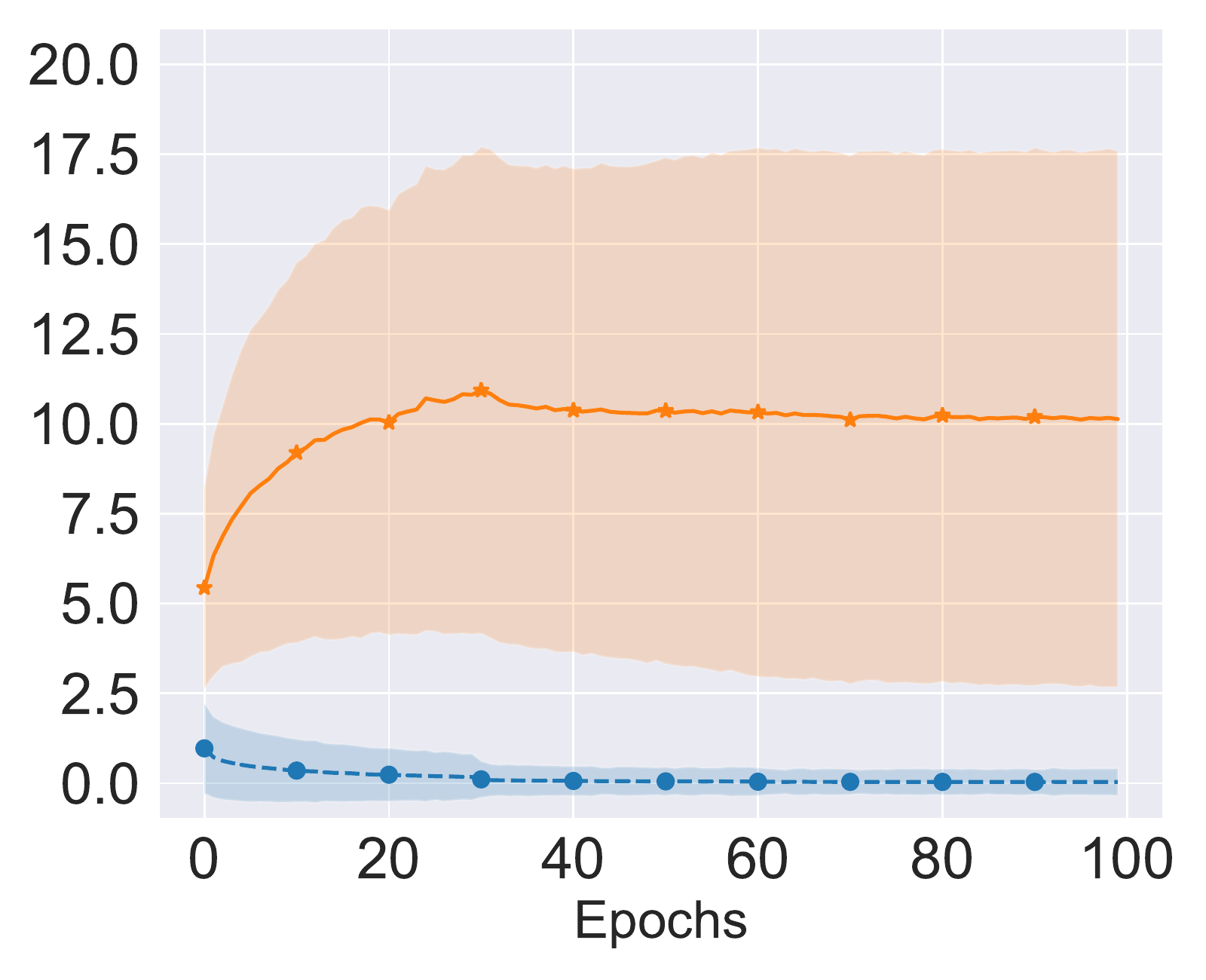}}
\subfigure[]{\label{fig:ce3}\includegraphics[width=0.245\textwidth]{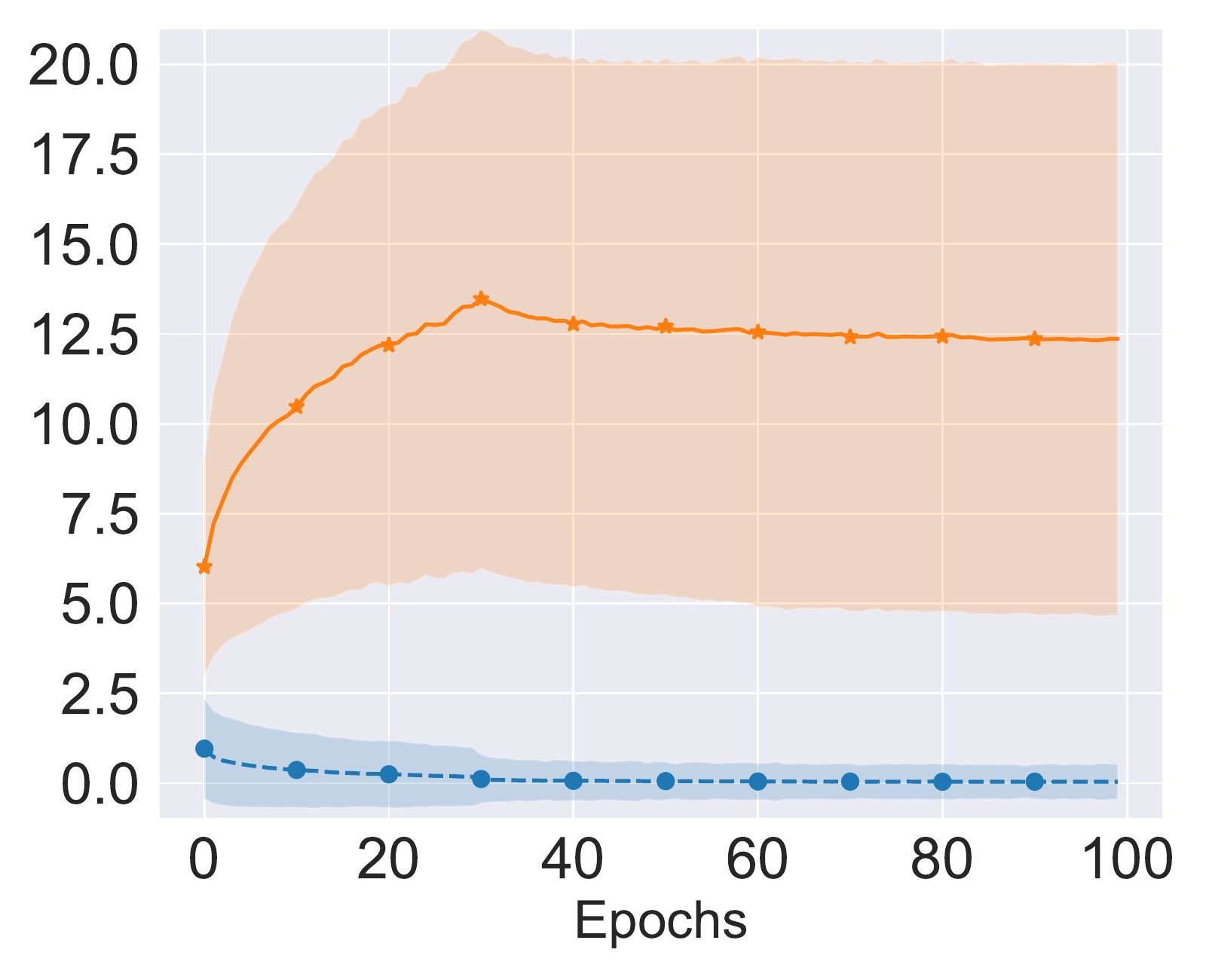}}
        % \caption{We trained the model using Importance Reweighting loss \cite{liu2015classification}, and the experiments setting is as same as Fig. \ref{fig:fig2}. We visualize the cross-entropy loss, weight $\hat{\beta}_i$ defined in Importance Reweighting, and reweighting loss. for better comparison, we experiment the different change of variance, \ie, penalizing, original, and increasing. The experiments results show that the $\hat{P}(Y|X)$ obtained by instances with incorrect labels is larger than correct one, and when increase the losses variance, the cross-entropy loss of instances with incorrect labels will increase, the $\hat{\beta}_i$ of these instances will decrease, and the $\hat{P}(Y|X)$ obtained by these instances will decrease(due to the decrease of $\hat{\beta}_i$).}\label{fig:ir}
            \caption{The change of losses with the increasing of training epochs for Reweighting.  (a) and (b) illustrate CE losses of $P(Y|X)$ without or with increasing variance, respectively.  (c) and (d) illustrate CE losses of $P(\tilde{Y}|X)$ without or with increasing variance, respectively.}\label{fig:ir}
\end{figure*}

\begin{figure*} [t] 
\centering
\subfigure[]{\label{fig:vol_clean2}\includegraphics[width=0.245\textwidth]{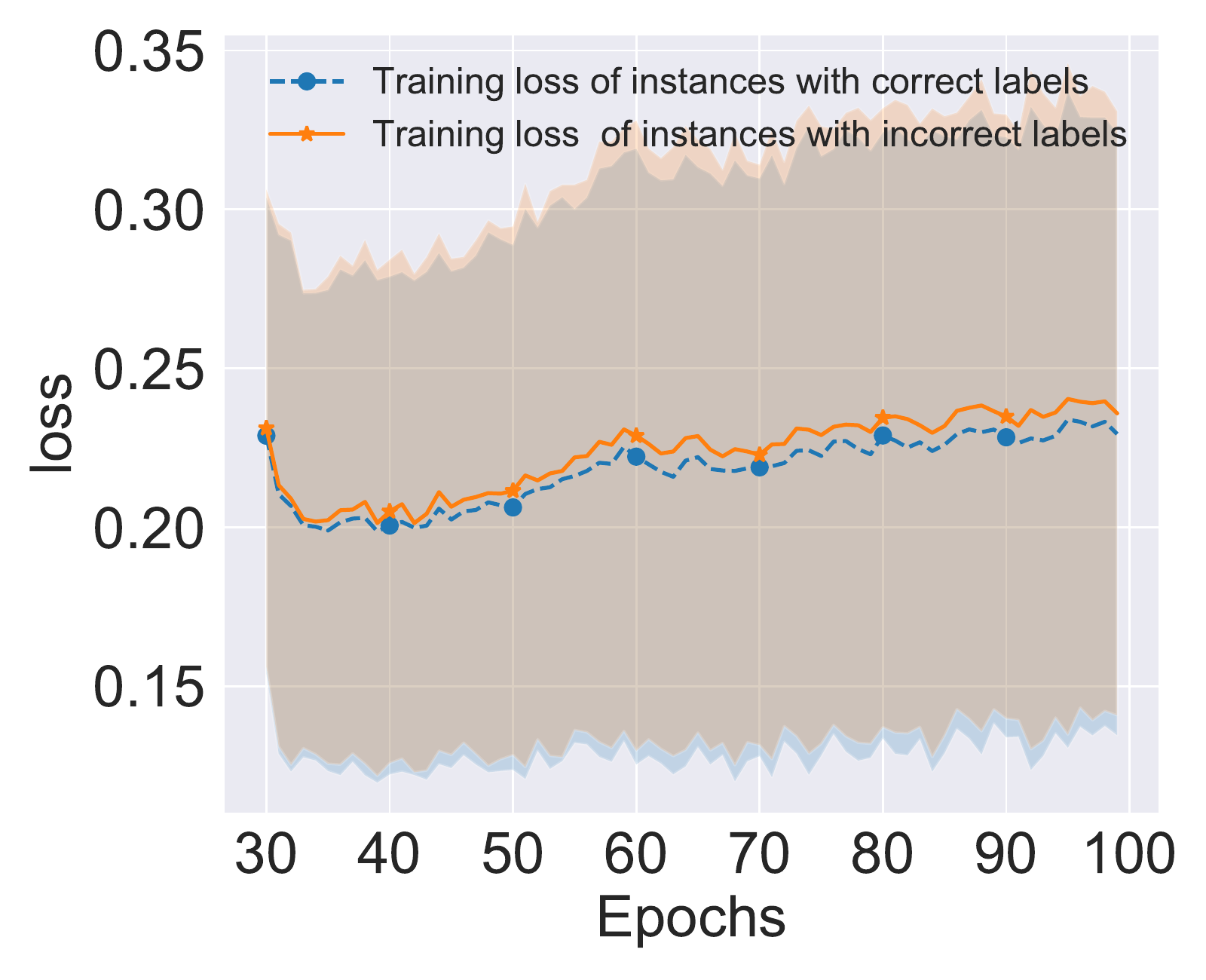}}
\subfigure[]{\label{fig:vol_clean3}\includegraphics[width=0.245\textwidth]{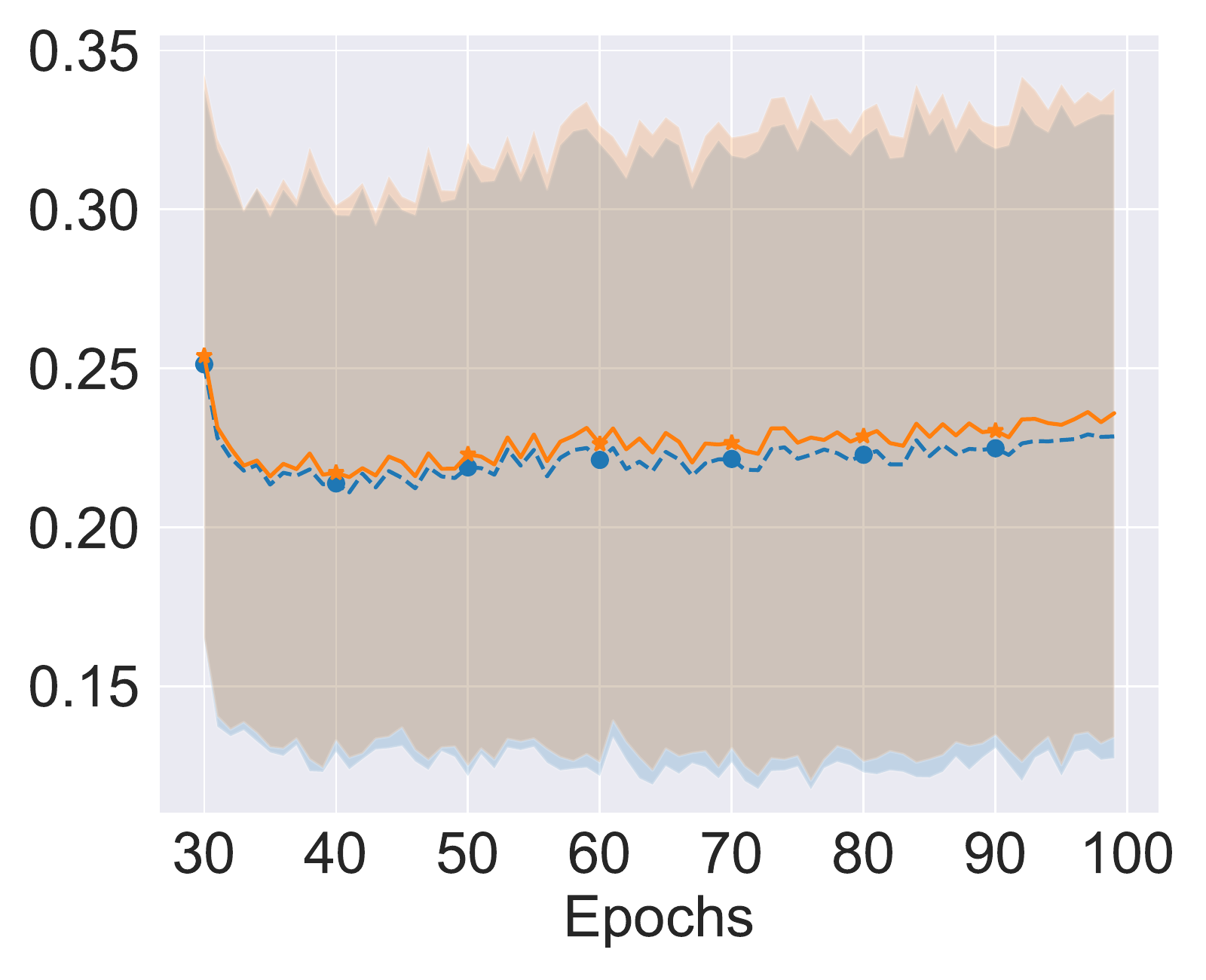}}
\subfigure[]{\label{fig:vol_noisy2}\includegraphics[width=0.245\textwidth]{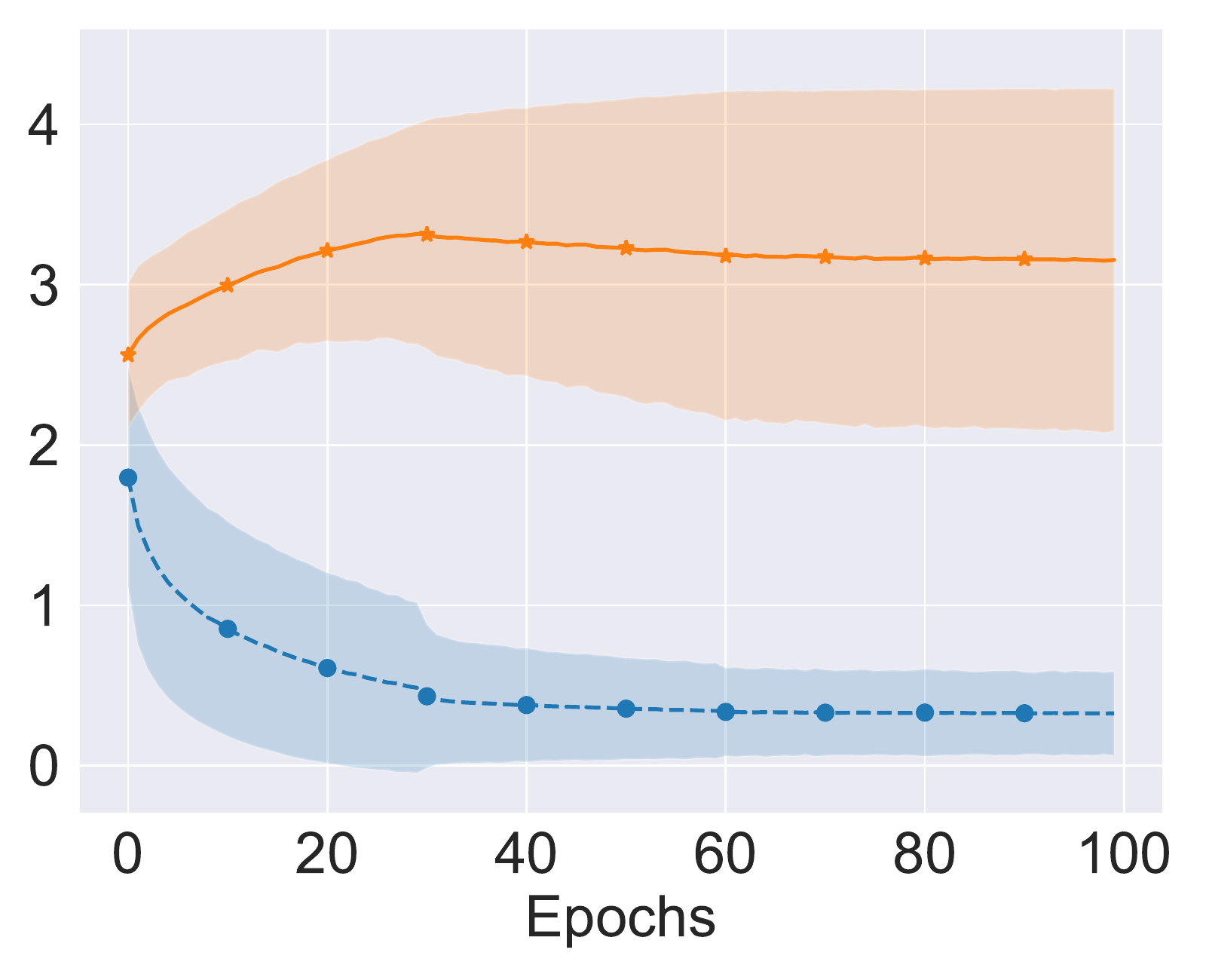}}
\subfigure[]{\label{fig:vol_noisy3}\includegraphics[width=0.245\textwidth]{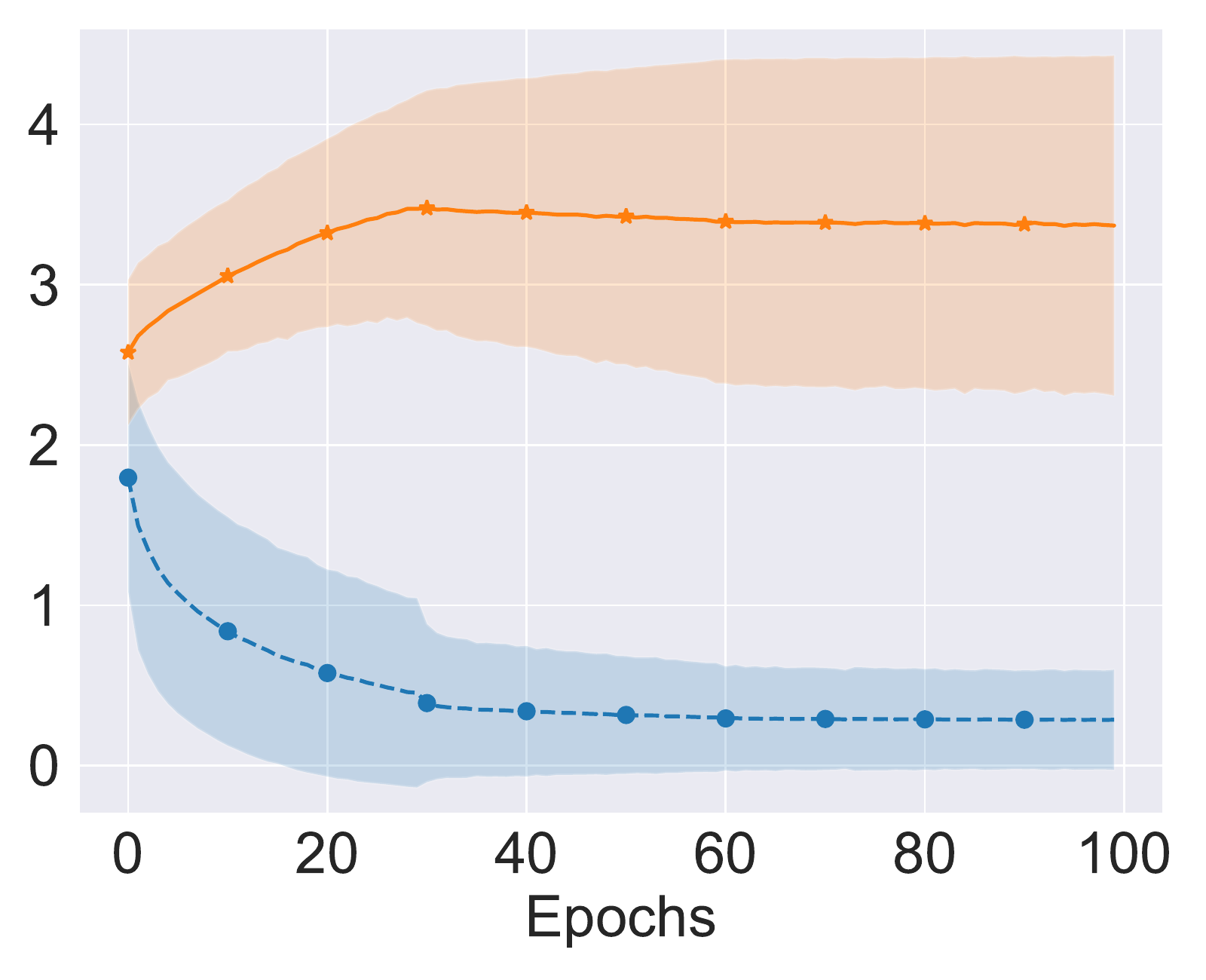}}

        \caption{The change of losses with the increasing of training epochs for VolMinNet.  (a) and (b) illustrate CE losses of $P(Y|X)$ without or with increasing variance, respectively.  (c) and (d) illustrate CE losses of $P(\tilde{Y}|X)$ without or with increasing variance, respectively.}\label{fig:vol}
\end{figure*}

\section{Experiments}\label{sec:exp}

\begin{table}[t]
\begin{center}
\resizebox{0.82\linewidth}{!}{
\begin{tabular}{ccccccc}
\toprule
 &\multicolumn{2}{c}{MNIST}&\multicolumn{2}{c}{CIFAR-10}&\multicolumn{2}{c}{CIFAR-100}\\
 & Sym-20\% & Sym-50\% & Sym-20\% & Sym-50\% & Sym-20\% & Sym-50\% \\
\midrule
Decoupling & 97.04 $\pm$ 0.06 & 94.58 $\pm$ 0.08 & 77.32 $\pm$ 0.35 & 54.07 $\pm$ 0.46 & 41.92 $\pm$ 0.49 & 22.63 $\pm$ 0.44 \\
MentorNet & 97.21 $\pm$ 0.06 & 95.56 $\pm$ 0.15 & 81.35 $\pm$ 0.23 & 73.47 $\pm$ 0.15 & 42.88 $\pm$ 0.41 & 32.66 $\pm$ 0.40 \\
Co-teaching & 97.07 $\pm$ 0.10 & 95.20 $\pm$ 0.23 & 82.27 $\pm$ 0.07 & 75.55 $\pm$ 0.07 & 48.48 $\pm$ 0.66 & 36.77 $\pm$ 0.52 \\
T-Revision & 98.72 $\pm$ 0.10 & 98.23 $\pm$ 0.10 & 87.95 $\pm$ 0.36 & 80.01 $\pm$ 0.62 & 62.72 $\pm$ 0.69 & 49.12 $\pm$ 0.22 \\
Dual T
 & 98.43 $\pm$ 0.05 & 98.15 $\pm$ 0.12 & 88.35 $\pm$ 0.33 & 82.54 $\pm$ 0.19 & 62.16 $\pm$ 0.58 & 52.49 $\pm$ 0.37 \\
\midrule
Forward & 97.47 $\pm$ 0.15 & $97.93\pm 0.22$ & $87.29\pm 0.63$ & $77.58\pm 1.05$ & $59.71\pm 0.40$ & $44.53\pm 1.11$ \\
Forward-VRNL & \textbf{98.89 $\pm$ 0.04$^*$} & \textbf{98.14 $\pm$ 0.27} & \textbf{89.81 $\pm$ 0.29$^*$} & \textbf{81.16 $\pm$ 0.55} & \textbf{68.19 $\pm$ 0.31$^*$} & \textbf{54.10 $\pm$ 1.2} \\
\midrule
Reweight & $98.20\pm 0.24$ & $97.93\pm 0.20$ & $88.42\pm 0.18$ & $82.13\pm 0.56 $ & $60.52\pm 0.52$ & $47.69\pm 0.78$ \\
Reweight-VRNL & \textbf{98.61 $\pm$ 0.21} & \textbf{98.27 $\pm$ 0.15$^*$} & \textbf{89.68 $\pm$ 0.24} & \textbf{83.99 $\pm$ 0.28$^*$} & \textbf{66.52 $\pm$ 0.25} & \textbf{50.26 $\pm$ 0.14} \\
\midrule
VolMinNet & $98.66\pm 0.14$ & $97.83\pm 0.15$ & $89.27\pm 0.30$ & $82.17\pm 0.19 $ & $65.65\pm 0.62$ & $54.40\pm 0.62$ \\
VolMinNet-VRNL & \textbf{98.78 $\pm$ 0.08} & \textbf{97.93 $\pm$ 0.20} & \textbf{89.42 $\pm$ 0.12} & \textbf{82.92 $\pm$ 0.24} & \textbf{66.40 $\pm$ 0.66} & \textbf{55.94 $\pm$ 0.64$^*$} \\
\midrule
%  &\multicolumn{2}{c}{MNIST}&\multicolumn{2}{c}{CIFAR-10}&\multicolumn{2}{c}{CIFAR-100}\\
 & Asym-20\% & Asym-50\% & Asym-20\% & Asym-50\% & Asym-20\% & Asym-50\% \\
 \midrule
Decoupling & 96.79 $\pm$ 0.01 & 94.71 $\pm$ 0.08 & 78.63 $\pm$ 0.27 & 71.01 $\pm$ 3.72 &  39.42 $\pm$ 0.48 & 21.64 $\pm$ 0.23 \\
MentorNet & 97.03 $\pm$ 0.05 & 94.66 $\pm$ 0.11 & 78.99 $\pm$ 0.34 & 68.00 $\pm$ 2.09 & 10.03 $\pm$ 0.33 & 11.14 $\pm$ 0.25 \\
Co-teaching & 97.02 $\pm$ 0.03 & 95.15 $\pm$ 0.09 & 83.96 $\pm$ 0.28 & 76.58 $\pm$ 0.84 & 13.36 $\pm$ 0.44 & 13.10 $\pm$ 0.66 \\
T-Revision & 98.90 $\pm$ 0.11 & \textbf{98.35 $\pm$ 0.13$^*$}  & 88.38 $\pm$ 0.56 & 81.51 $\pm$ 0.74 & 59.52 $\pm$ 0.43 & 45.56 $\pm$ 1.86 \\
Dual T & 95.46 $\pm$ 0.14 & 91.46 $\pm$ 0.29 & 70.31 $\pm$ 1.06 & 53.04 $\pm$ 2.76 & 05.80 $\pm$ 0.78 & 02.38 $\pm$ 0.94 \\
\midrule
Forward & 98.42 $\pm$ 0.06 & $97.92\pm 0.04$ & $87.70\pm 0.29$ & $79.25\pm 1.61$ & $60.24\pm 0.42$ & $43.39\pm 1.15$ \\
Forward-VRNL & \textbf{98.92 $\pm$ 0.06$^*$} & \textbf{98.13 $\pm$ 0.21} & \textbf{89.98 $\pm$ 0.11$^*$} & \textbf{82.35 $\pm$ 0.88} & \textbf{67.89 $\pm$ 0.30$^*$} & \textbf{53.67 $\pm$ 0.52} \\
\midrule
Reweight & $98.50\pm 0.07$ & $98.09\pm 0.08$ & $88.55\pm 0.32$ & $82.72\pm 0.38 $ & $60.81\pm 0.70$ & $46.36\pm 0.18$ \\
Reweight-VRNL & \textbf{98.77 $\pm$ 0.21} & \textbf{98.10 $\pm$ 0.16} & \textbf{89.80 $\pm$ 0.11} & \textbf{84.20 $\pm$ 0.25$^*$} & \textbf{66.62 $\pm$ 0.45} & \textbf{49.71 $\pm$ 0.64} \\
\midrule
VolMinNet & 98.62 $\pm$ 0.10 & 98.03 $\pm$ 0.12 & 89.50 $\pm$ 0.18 & 83.15 $\pm$ 0.56 & 66.02 $\pm$ 0.73 & 55.17 $\pm$ 0.46 \\
VolMinNet-VRNL & \textbf{98.76 $\pm$ 0.13} & \textbf{98.08 $\pm$ 0.08} & \textbf{89.64 $\pm$ 0.19} & \textbf{83.65 $\pm$ 0.32} & \textbf{66.24 $\pm$ 0.95} & \textbf{55.85 $\pm$ 0.73$^*$} \\
\midrule
%  &\multicolumn{2}{c}{MNIST}&\multicolumn{2}{c}{CIFAR-10}&\multicolumn{2}{c}{CIFAR-100}\\
 & Pair-20\% & Pair-45\% & Pair-20\% & Pair-45\% & Pair-20\% & Pair-45\% \\
 \midrule
Decoupling & 96.93 $\pm$ 0.07  & 94.34 $\pm$ 0.54 & 77.12 $\pm$ 0.30 & 53.71 $\pm$ 0.99 & 40.12 $\pm$ 0.26  & 27.97 $\pm$ 0.12 \\
MentorNet & 96.89 $\pm$ 0.04 & 91.98 $\pm$ 0.46 & 77.42 $\pm$ 0.23 & 61.03 $\pm$ 0.20 & 39.22 $\pm$ 0.47 & 26.48 $\pm$ 0.37 \\
Co-teaching & 97.00 $\pm$ 0.06 & 96.25 $\pm$ 0.01 & 80.65 $\pm$ 0.20 & 73.02 $\pm$ 0.23 & 42.79 $\pm$ 0.79 & 27.97 $\pm$ 0.20 \\
T-Revision & 98.89 $\pm$ 0.08 & 84.56 $\pm$ 8.18 & 90.33 $\pm$ 0.52 & 78.94 $\pm$ 2.58 & 64.33 $\pm$ 0.49 & 41.55 $\pm$ 0.95 \\
Dual T & 98.86 $\pm$ 0.04 & 96.71 $\pm$ 0.12 & 89.77 $\pm$ 0.25 & 76.53 $\pm$ 2.51 & 67.21 $\pm$ 0.43 & 47.60 $\pm$ 0.43 \\
\midrule
Forward & 98.85 $\pm$ 0.09 & 96.45 $\pm$ 4.03 & \textbf{90.88 $\pm$ 0.23$^*$} & 83.27 $\pm$ 9.47 & 62.54 $\pm$ 0.42 & 41.96 $\pm$ 1.45 \\
Forward-VRNL & \textbf{98.88 $\pm$ 0.08} & \textbf{96.55 $\pm$ 3.88} & \textbf{90.88 $\pm$ 0.29$^*$} & \textbf{83.54 $\pm$ 9.29} & \textbf{62.78 $\pm$ 0.32} & \textbf{42.29 $\pm$ 1.23} \\
\midrule
Reweight & 98.64 $\pm$ 0.07 & 95.52 $\pm$ 3.58 & 89.85 $\pm$ 0.30 & 76.03 $\pm$ 5.02 & 61.35 $\pm$ 0.66  & 40.10 $\pm$ 0.46 \\
Reweight-VRNL & \textbf{98.68 $\pm$ 0.14} & \textbf{95.97 $\pm$ 3.52} & \textbf{89.77 $\pm$ 0.39} & \textbf{76.75 $\pm$ 6.15} & \textbf{61.37 $\pm$ 0.42} & \textbf{40.30 $\pm$ 0.57} \\
\midrule
VolMinNet & \textbf{99.05 $\pm$ 0.05$^*$} & 99.08 $\pm$ 0.06 & 90.73 $\pm$ 0.23 & 88.47 $\pm$ 0.61 & 69.96 $\pm$ 1.18 & 61.85 $\pm$ 1.41 \\
VolMinNet-VRNL & 99.02 $\pm$ 0.08 & \textbf{99.10 $\pm$ 0.08$^*$} & \textbf{90.86 $\pm$ 0.27} & \textbf{88.77 $\pm$ 0.51$^*$} & \textbf{70.18 $\pm$ 0.50$^*$} & \textbf{63.38 $\pm$ 1.72$^*$} \\
\bottomrule
\end{tabular}
}
\caption{Means and standard deviations (percentage) of classification accuracy. Results with ``*'' mean that they are the highest accuracy.}
\label{tb:sym_result}
\end{center}
\end{table}

In this section, we first illustrate the empirical performance of VRNL and other baselines on both synthetic and real-world noisy datasets. What is more, we will delve into the different effects of the proposed method on correct and incorrect samples to verify its effectiveness on Section \ref{sec:var}. Finally, we will illustrate the robustness of the proposed method when the estimated transition matrix is biased.

\begin{table*}
  \centering
  \resizebox{0.85\linewidth}{!}{
  \begin{tabular}{cccccc}
    \toprule
    Decoupling & MentorNet & Co-teaching & T-Revision & Dual T & PTD \\
    \midrule
    54.53 & 56.79 & 60.15 & 70.97 & 70.17 & 71.67 \\
    \midrule
    Forward & Forward-VRNL & Reweight & Reweight-VRNL & VolMinNet & VolMinNet-VRNL \\
    71.27 & \textbf{72.43} & 71.62 & \textbf{72.14} & 72.29 & \textbf{72.66} \\
    \bottomrule
  \end{tabular}
  }
  \caption{Classification accuracy(percentage) on Clothing1M. Only noisy data are exploited for training and validation.}
  \label{tb:clothing}
\end{table*}
\begin{figure*} [t] 
\centering
        \subfigure[Reweight on MNIST]{\label{fig:fig3a}\includegraphics[width=0.245\textwidth]{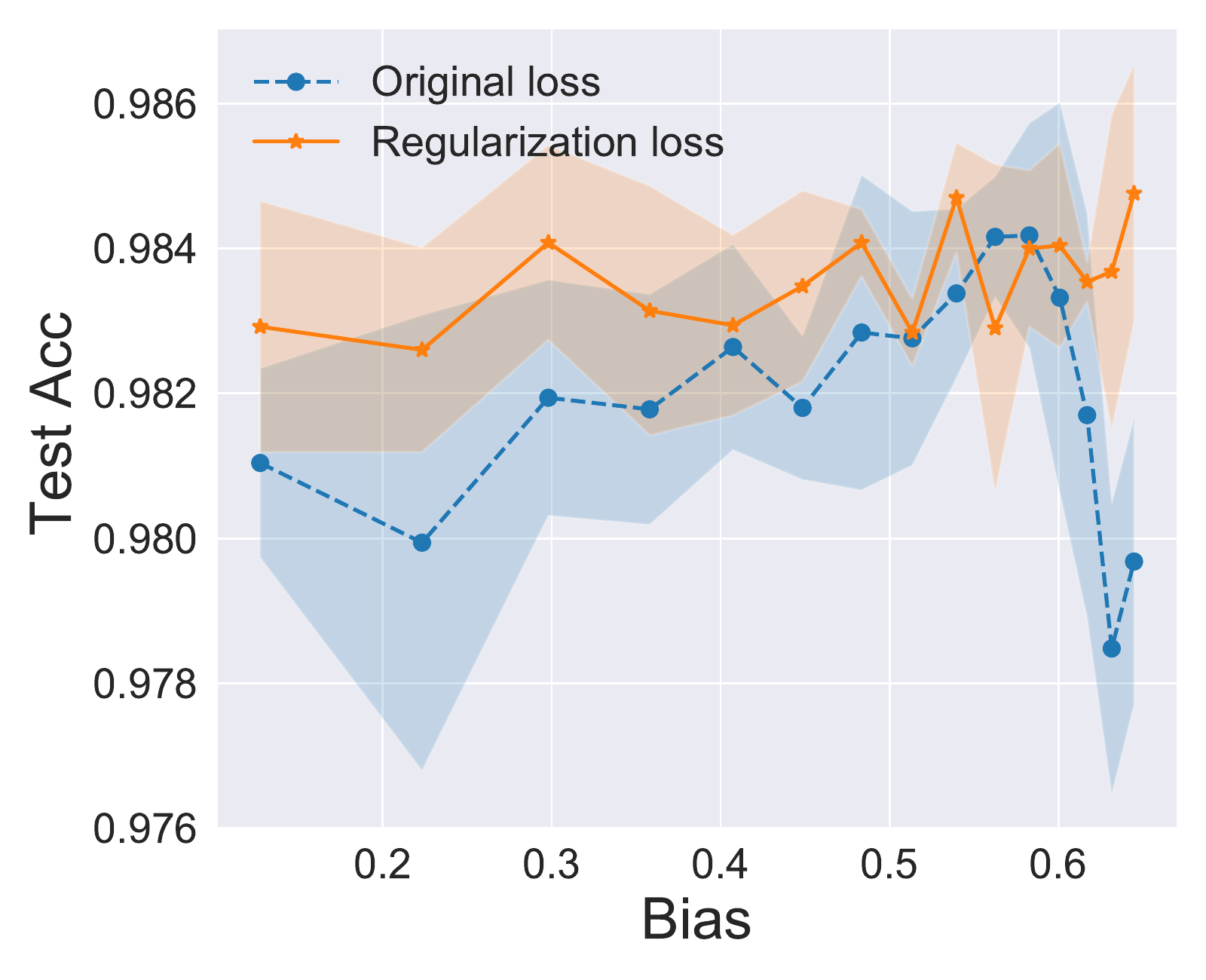}}
        \subfigure[Forward on MNIST]{\label{fig:fig3b}\includegraphics[width=0.245\textwidth]{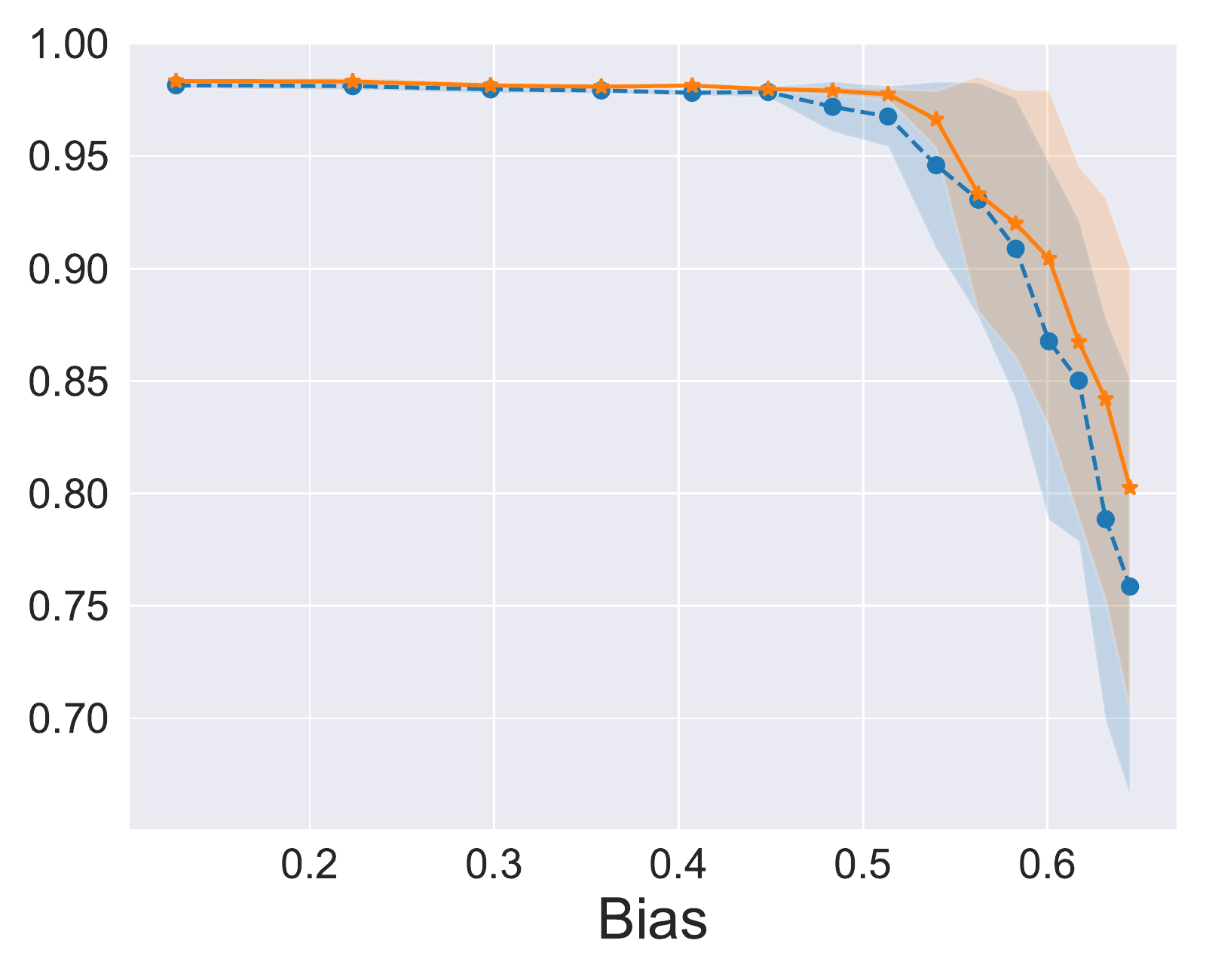}}
        \subfigure[Reweight on CIFAR10]{\label{fig:fig3c}\includegraphics[width=0.245\textwidth]{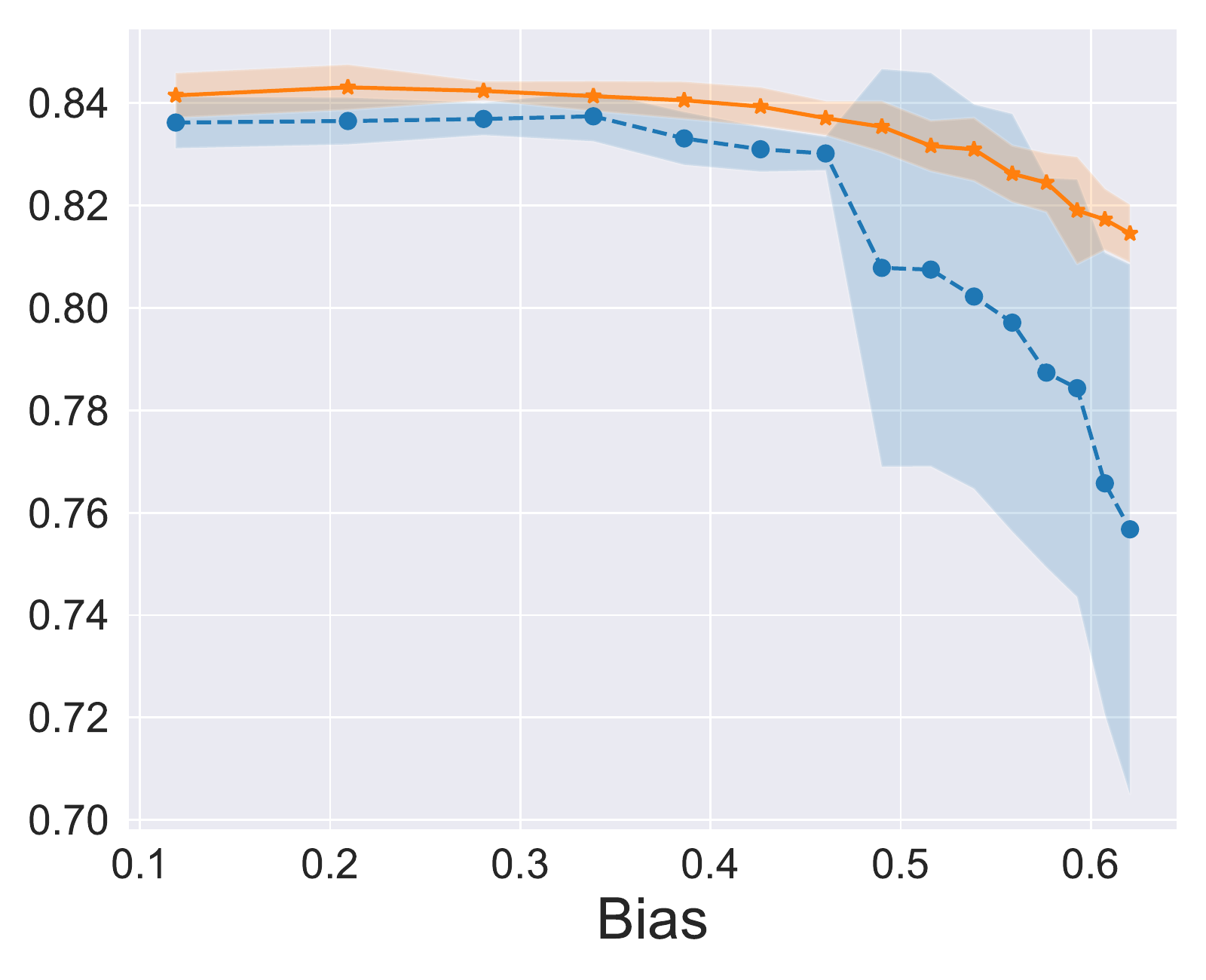}}
        \subfigure[Forward on CIFAR10]{\label{fig:fig3d}\includegraphics[width=0.245\textwidth]{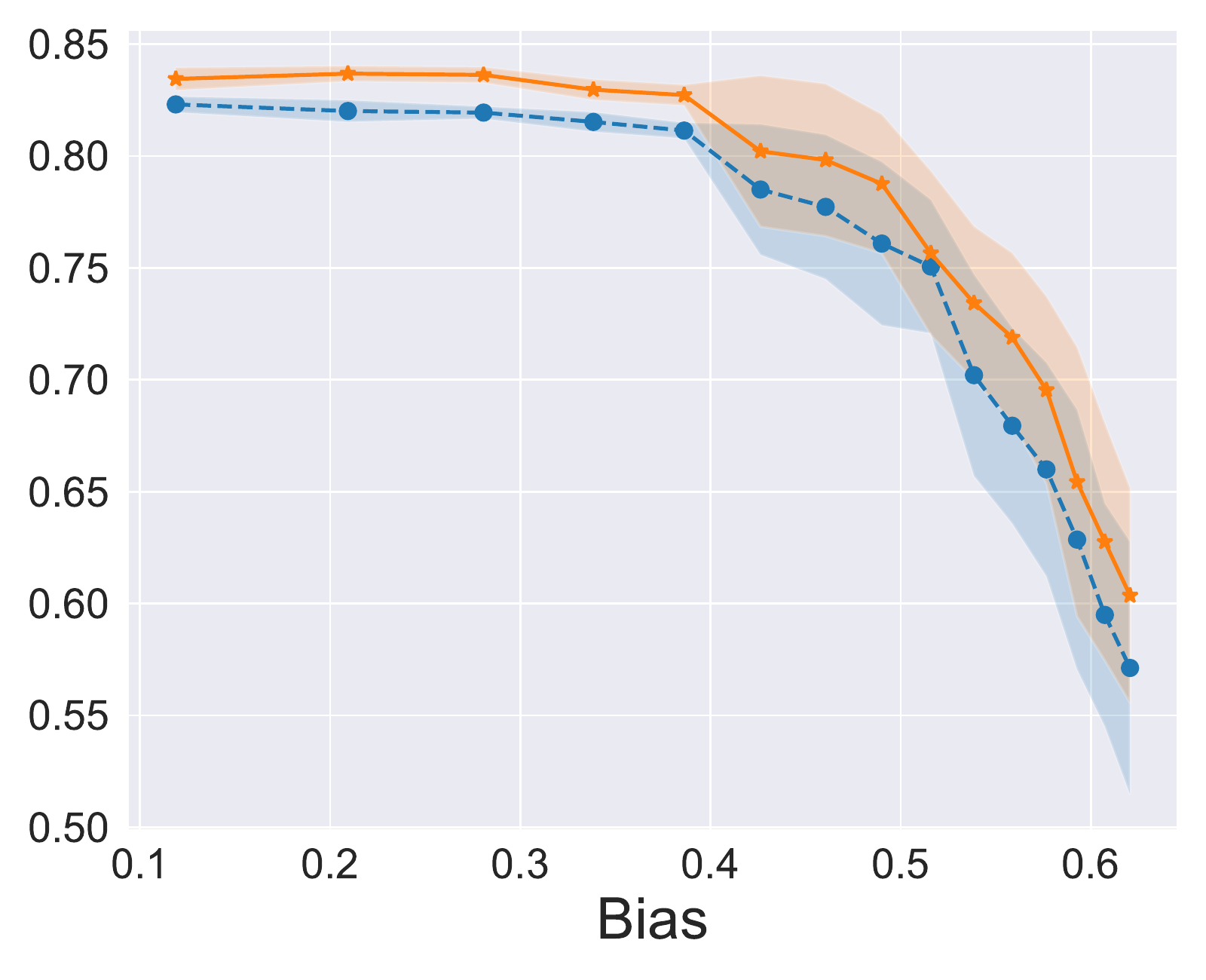}}
        \caption{Test accuracies of the models trained on MNIST and CIFAR10 by using biased transition matrices, We increase the error of transition matrices manually. The proposed VRNL is robust to the biased transition matrix.}\label{fig:fig3}    
\end{figure*}

\textbf{Datasets.} We verify the performance of proposed method on the manually corrupted version of three datasets, \ie, MNIST \cite{lecun2010mnist}, CIFAR-10 \cite{krizhevsky2009learning} and CIFAR-100 \cite{krizhevsky2009learning}, and one real-world noisy dataset, \ie, Clothing1M \cite{xiao2015learning}. We leave out 10\% of training data as validation sets. The experiments are repeated five times on the synthetic noisy datasets. Clothing1M \cite{xiao2015learning} contains 1M images with real-world noisy labels, it also contains 50k, 14k, and 10k images with clean labels for training, validation, and testing, respectively. Existing methods like Forward \cite{patrini2017making} and T-revision \cite{xia2019anchor} use the 50k clean data to initialize the transition matrix and validate on 14k clean data. We assume that the clean data is not accessible, and therefore, the clean data are not used for training and validation. We leave out 10\% of examples from 1M noisy data for validation.

\textbf{Baselines.}
The baselines used in our experiments: 1). Decoupling \cite{malach2017decoupling} trains two models at the same time, and only the instances which have different predictions from two networks are used to update the parameter; 2). MentorNet \cite{jiang2018mentornet} pre-trains an extra model which is used to select clean examples for the main model training; 3). Co-teaching \cite{han2018co} trains two networks simultaneously, and each network is used to select small-loss examples as trust examples to its peer network for further training; 
4). Forward \cite{patrini2017making} 
estimates the transition matrix in advance, then uses it to approximate the clean class posteriors;
5). T-Revision \cite{xia2019anchor} proposes a method to fine-tune the estimated transition matrix to improve the classification performance; 6). Dual T \cite{yao2020dual} improves the estimation of the transition matrix by introducing an intermediate class, and then factorizes the transition matrix into the product of two easy-to-estimate transition matrices; 7). VolMinNet \cite{li2021provably} is an end-to-end label-noise learning method, which can learn the transition matrix and the classifier simultaneously;
8). Reweight \cite{liu2015classification} uses the importance reweighting technique to estimate the expected risk on the clean domain by using noisy data.  

\textbf{Noise Types.} 
To generate a noisy dataset, we corrupted the training and validation sets manually according to a special transition matrix $\bm{T}$. Specifically, we conduct experiments on synthetic noisy datasets with three wildly used types of noise: 1).  Symmetry flipping (Sym-$\epsilon$) \cite{patrini2017making}; 2).  Asymmetry flipping (Asym-$\epsilon$); 3).  Pair flipping (Pair-$\epsilon$) \cite{han2018co}. We manually corrupt the labels of instances according to the transition matrix $\bm{T}$.
% For example, for the noise type Sym-$50\%$, we corrupt an instance with label $i$ to other categories $\tilde{y}$, where $\tilde{y} \in \{1,2,\dots, C\}$, by sampling a number from the distribution of i-th column of $\bm{T}$, and it leads around 50\% of instances to have label noise. Different type of noise has different transition matrix, therefore the distribution for generating spotted labels is different.

\textbf{Network structure and optimization.} 
We implement the proposed methods and baseline using Pytorch 1.9.1 and train the models on TITAN Xp. The model structure and optimizer are as same as the state-of-the-art method \cite{li2021provably}. Specifically, we use a LeNet-5 network \cite{lecun1998gradient} for MNIST, a ResNet-18 network for CIFAR-10, a ResNet-32 network \cite{he2016deep} for CIFAR-100, a ResNet-50 pretrained on ImageNet for Clothing 1M.
On synthetic noise datasets, SGD is used to train the neural network with batch size $128$, momentum $0.9$, weight decay $10^{-4}$, and an initial learning rate $10^{-2}$. The algorithm is trained for $80$ epochs, and the learning rate is divided by 10 after the $30$-th and $60$-th epochs. 
For Forward and Reweight, we set the hyper-parameter $\alpha=0.1$ on symmetry-flipping noise, asymmetry-flipping noise, and we set $\alpha=0.01$ on pair-flipping noise.
For VolMinNet, we set $\alpha=0.005$ on MNIST and CIFAR-100 with pair-flipping noise.
For other experiments on synthetic noisy datasets, $\alpha=0.05$ is employed.
When the dataset is Clothing-1M, for Forward and Reweight, SGD with batch size 64, momentum 0.9, weight decay $10^{-4}$ is used to train the model, and $\alpha$ is set to be $0.1$; for VolMinNet, SGD with batch size 64, momentum 0.9, weight decay $10^{-3}$ is used, and $\alpha$ is set to be $0.005$.
For Forward and Reweight, the transition matrix $\bm{T}$ has to be estimated in advance.
For the end-to-end method VolMinNet, the transition matrix $\bm{T}$ and the classifier are learned simultaneously. To estimate the transition matrix, we follow the same experimental settings described in their original papers \cite{patrini2017making,li2021provably}.

\subsection{Classification Accuracy Evaluation}

We embed VRNL into the label-noise learning methods, \eg, Forward, Reweight and VolMinNet which are named Forward-VRNL, Reweight-VRNL and VolMinNet-VRNL, respectively. In Table \ref{tb:sym_result}, we illustrate classification accuracies on datasets containing symmetry-flipping, asymmetry-flipping and pair-flipping noise. It shows that VRNL improves the classification accuracies of all the label-noise learning methods on different datasets and different types of noise.
It is worth noting that for Forward and Reweight, the estimated transition matrices usually have larger estimated errors compared with VolMinNet \cite{li2021provably}. The performance of VolMinNet is usually better than that of Forward and Reweight. However, by employing our method, the performance of Forward and Reweight are comparable to that of VolMinNet, which suggests that VRNL is robust to the biased transition matrix.

In Table \ref{tb:clothing}, we illustrate the results on the real-world dataset Clothing1M. The proposed method improves the generalization ability of backbone methods. The performance of VolMinNet-VRNL outperforms all other baselines.

\subsection{The influence on clean and noisy class posteriors}

To analyze the influence of variance increase on clean class posteriors and clean class posteriors. In Fig.~\ref{fig:ir} and Fig.~\ref{fig:vol}, we visualize the change of cross-entropy losses for instances with clean labels and instances with noisy labels during the model training, respectively.
The average loss and the standard derivation of mislabeled examples and correctly labeled examples are visualized separately for better illustration. The methods used are Reweight and VolMinNet.

By comparing Reweight-VRNL with Reweight, the loss of noisy labels for mislabeled examples is larger but the loss of noisy labels for correctly labeled examples is almost unchanged as shown in Fig. \ref{fig:ce2} and Fig. \ref{fig:ce3}. It means that the proposed method prevents the model from memorizing incorrect labels and has little influence on learning correctly labeled examples.
By comparing Fig. \ref{fig:ir_clean3} with Fig. \ref{fig:ir_clean2}, the loss of clean labels for mislabeled examples becomes smaller when our method is employed. It implies that our method helps learn clean class posteriors of mislabeled examples. 

Similarly, the results also hold for VolMinNet. 
Specifically, the variance of noisy class posteriors in Fig. \ref{fig:vol_noisy3} increases compared with Fig. \ref{fig:vol_noisy2}, which could help VolMinNet better estimate the transition matrix. It is because our method encourages the diversity of noisy class posteriors, which makes the \textit{sufficiently scattered} assumption easier to be satisfied when the sample size is limited. Meanwhile, it can be seen that the empirical risk defined by clean clean training samples decreases after using our method, as shown in Fig. \ref{fig:vol_clean2} and \ref{fig:vol_clean3}. It means that the model can classify the examples better.

\subsection{Performance with the biased Transition Matrix}
\label{anti}

In practice, the noise transition matrix generally is not given and is required to be estimated. However, the estimated transition matrix could contain a large estimation error. One reason is that the transition matrix can be hard to accurately estimate when sample size is limited \cite{yao2020dual}. Another reason is that the assumptions \cite{patrini2017making,li2021provably} used to identify the transition matrix may not be held. This motivates us to investigate the performance of our regularizer when the transition matrix contains bias. 

To simulate the estimation error, we manually inject noise into the transition matrix, i.e.,
\begin{align}
\bm{T}^{\rho}=\bm{T}+\gamma | \Delta | \nonumber,
\end{align}
where $\Delta \in \mathbb{R}^{ C\times C}$ sampled from standard multivariate normal distribution, and $\gamma \in [0.01, 0.15]$. Then we normalize the column of the transition matrix $\bm{T}_{\rho}$ sum up to 1 by
\begin{align}
\bm{T}^N_{ij}=\frac{{\bm{T}^{\rho}}_{ij}}{\sum\limits_{k=1}^C {\bm{T}^{\rho}}_{ik}} \nonumber.
\end{align}
The estimation error $\epsilon_T$ of a transition matrix is calculated by employing the entry-wise matrix norm, i.e.,
\begin{align}
\epsilon_{\bm{T}} =\frac{\Vert \bm{T}-\bm{T}^N\Vert _{1,1}}{\Vert \bm{T}\Vert _{1,1}}.\nonumber
\end{align}

The biased transition matrix $T^N$ is adopted to Reweight, Reweight-VRNL, Forward and Forward-VRNL, respectively. Experimental results shown in Fig. \ref{fig:fig3} illustrate that our method is more robust to the bias transition matrix. 
Specifically, for most experiments and different levels of bias $\epsilon_T$, the test accuracies of Reweight-VRNL and Forward-VRNL are higher than Reweight and Forward. 
Additionally, the test accuracy of Reweight-VRNL drops much slower than Reweight with the increasing of bias $\epsilon_T$.

\section{Conclusion}

In this paper, we study whether we should penalize the variance of losses for the problem of learning with noisy labels. Interestingly, we found that the variance of losses should be increased, which can boost the memorization effects and reduce the harmfulness of incorrect labels. 
Theoretically, we show that increasing variance of losses can reduce the weights of the gradient with respect to instances with incorrect labels, therefore these instances have a small contribution to the update of model parameters.
A simple and effective method VRNL is also proposed which can be easily integrated into existing label-noise learning methods to improve their robustness.
The experimental results on both synthetic and real-world noisy datasets demonstrate that the proposed VRNL can dramatically improve the performance of existing label-noise learning methods.
Empirically, we have shown that the proposed method can help models better learn clean class posteriors.
We have also illustrated that VRNL can improve the classification performance of existing methods when the transition matrix is poorly estimated, which makes our method be more practically useful.

\bibliographystyle{plainnat}
\bibliography{bib}

\end{document}